\documentclass[12pt,review]{elsarticle}
\usepackage{multirow} 
\usepackage{lineno,hyperref}
\modulolinenumbers[5]
\usepackage{subcaption}
\usepackage{xcolor,soul}
\usepackage{blindtext}
\usepackage{booktabs}
\usepackage{amsbsy} 
\usepackage{upgreek}
\usepackage{amsmath}
\usepackage{bbold}

\usepackage[margin=3cm]{geometry}
\begin{document}
 
\begin{frontmatter}


\title{A Data-Driven Approach to Full-Field Damage and Failure Pattern Prediction in Microstructure-Dependent Composites using Deep Learning}

\author[firstadress,secondaddress]{Reza Sepasdar}
\author[secondaddress]{Anuj Karpatne}
\author[firstadress]{Maryam Shakiba\corref{mycorrespondingauthor}}
\cortext[mycorrespondingauthor]{Corresponding author}
\ead{mshakiba@vt.edu}

\address[firstadress]{Department of Civil and Environmental Engineering, Virginia Tech, 750 Drillfield Dr., Blacksburg, VA 24061, USA}
\address[secondaddress]{Department of Computer Science, Virginia Tech, 225 Stanger Street, Blacksburg, VA 24061, USA}

\begin{abstract}
An image-based deep learning framework is developed in this paper to predict damage and failure in microstructure-dependent composite materials. The work is motivated by the complexity and computational cost of high-fidelity simulations of such materials. The proposed deep learning framework predicts the post-failure full-field stress distribution and crack pattern in two-dimensional representations of the composites based on the geometry of microstructures. The material of interest is selected to be a high-performance unidirectional carbon fiber-reinforced polymer composite. The deep learning framework contains two stacked fully-convolutional networks, namely, Generator 1 and Generator 2, trained sequentially. First, Generator 1 learns to translate the microstructural geometry to the full-field post-failure stress distribution. Then, Generator 2 learns to translate the output of Generator 1 to the failure pattern. A physics-informed loss function is also designed and incorporated to further improve the performance of the proposed framework and facilitate the validation process. In order to provide a sufficiently large data set for training and validating the deep learning framework, 4500 microstructural representations are synthetically generated and simulated in an efficient finite element framework. It is shown that the proposed deep learning approach can effectively predict the composites' post-failure full-field stress distribution and failure pattern, two of the most complex phenomena to simulate in computational solid mechanics.
\end{abstract}



\end{frontmatter}


\section{Introduction}

The mechanical behavior of composite materials depends on the geometry of their microstructures. Hence, accurate analysis of the overall mechanical behavior, specifically when nonlinearity and damage are present, requires numerical simulation on the microstructure level. In the simulations, a representative volume element (RVE) of the microstructure is studied, typically in a static displacement-controlled scheme.

Finite element (FE) analysis is widely being used for conducting such simulations. FE simulations, however, have a number of shortcomings:
\begin{enumerate}
    \item A robust FE simulation requires that accurate constitutive equations to be considered for different RVE constituents. The constitutive equations to model damage and failure of the materials are complex and their implementation is not straightforward. 
    \item The complex geometry of heterogeneous materials microstructure necessitates the application of a fine mesh, involving many finite elements, to the numerical model. Hence, several degrees of freedom are present, demanding large amounts of memory.
    \item In FE simulations of damage and failure, loading has to be applied in several increments. For each increment, a system of equations has to be solved iteratively to calculate the nodal displacements based on the internal force vector and stiffness matrix. Each iteration contains two nested loops, one over all the elements and another over the integration points to calculate and assemble the global stiffness matrix and internal force vector. This procedure renders the FE simulations time-consuming and highly computationally expensive.
    \item Simulating damage propagation and failure of materials is followed by numerical instabilities, resulting in several time-step refinements by the solver. Hence, several additional loading increments are required for the completion of the simulations, imposing addition computational cost. 
\end{enumerate}

Hence, FE simulations are complex, time-consuming, and demand significant computational power. As a result, there has been an increasing interest in using deep learning approaches as surrogates for the expensive FE simulations in the past few years. While there have been major advances in the application of deep learning to computational fluid mechanics (e.g., \cite{raissi2019deep,muralidhar2020physics,raissi2020hidden,kashefi2021point}), its application to the computational solid mechanics area is still at an early stage of research. 

In computational solid mechanics, deep learning techniques have been mainly employed to predict different global mechanical attributes of heterogeneous materials based on the microstructural geometry, e.g., homogenized elastic properties \cite{cecen2018material,yang2018deep,vlassis2020geometric,herriott2020predicting} and inelastic material response \cite{wang2018multiscale,mozaffar2019deep,heider2020so}. These attributes are presented by a few scalars, describing the macroscopic response of a heterogeneous material under certain loading scenarios.

A few researchers used deep learning to predict full-field microscopic attributes of materials with simple geometries. Donegan et al. \cite{donegan2019associating} used convolutional neural networks to identify the locations of thermally-induced stress hot spots in particulate composites based on an image of the microstructure. Fend and Prabhakar \cite{feng2020difference} proposed a difference-based neural network framework to predict the elastic stress distribution around a single fiber, affected by the presence of four uniformly distributed fibers and the boundaries. A few research works recently utilized physics-informed deep learning approaches to predict stress distribution and displacement field \cite{zhang2020physics,abueidda2020deep,haghighat2021sciann} in solid mechanics problems. Learning in an unsupervised fashion is an important advantage of the physics-based models. However, the implementation of these methods is complex as they rely on designing an appropriate objective function based on the constitutive equations and mechanical laws, governing that specific problem.
As a result, the application of deep learning to predict the full-field stress distribution in heterogeneous materials with complex microstructures in the presence of damage has not been attempted.

Besides, the application of deep learning approaches to predict failure pattern of materials and structures under mechanical loading is quite scarce in the literature, while the majority of research efforts have been allocated to the detection and classification of failure and damage (e.g., \cite{dong2020microstructural,pazdernik2020microstructural,lei2020lost,mao2020toward,soleimani2021system}). Hunter et al. \cite{hunter2019reduced} used artificial neural networks with a simple architecture to predict crack pattern in brittle concrete wall specimens, subjected to tension. The macrostructure of the specimens contained 20 pre-existing cracks with the same length and random orientations and locations. Their proposed model was trained based on a number of predefined geometrical features as the significant predictors of the crack path. Schwarzer et al. \cite{schwarzer2019learning} used convolutional and recurrent neural networks to predict the evolution of the crack pattern in concrete wall specimens with pre-existing cracks, similar to those from Hunter et al.'s study  \cite{hunter2019reduced}. Pierson et al. \cite{pierson2019predicting} proposed a convolutional neural networks framework to predict the roughness of crack surface in a polycrystalline alloy based on the microstructure. The method, however, requires the prior knowledge of the crack location and can only predict the crack surface topography. 
As a result, the existing research works on failure pattern prediction problems studied only simple scenarios. The prediction of failure pattern in heterogeneous materials with complex microstructures using deep learning was not studied before.

In summary, utilizing deep learning techniques to predict full-field stress distribution in the presence of damage as well as failure pattern for heterogeneous materials with complex microstructures were not attempted before. Hence, this work proposes an end-to-end deep learning framework to predict post-failure stress distribution and failure pattern within two-dimensional (2-D) representations of microstructure-dependent composites. The predictions are based on an image of the microstructure as the input, making the framework convenient to use.

\section{Overview of the Proposed Deep Learning Framework} \label{section: overview}
\begin{figure*}
    \centering
	\begin{subfigure}{0.3\textwidth}
	    \centering
        \includegraphics[width=\textwidth]{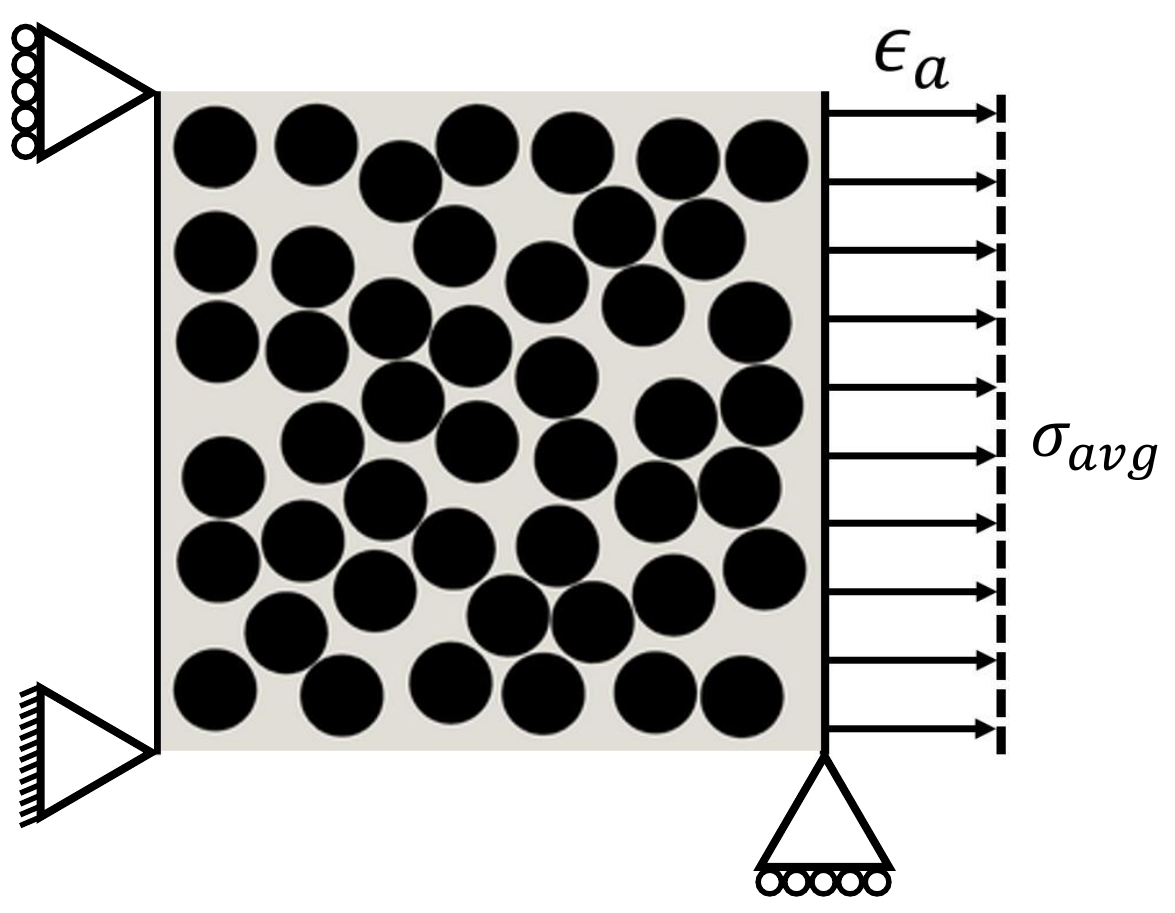}
        \caption{} \label{fig:boundary_condition}
    \end{subfigure}
    \begin{subfigure}{0.4\textwidth}
	    \centering
        \includegraphics[width=\textwidth]{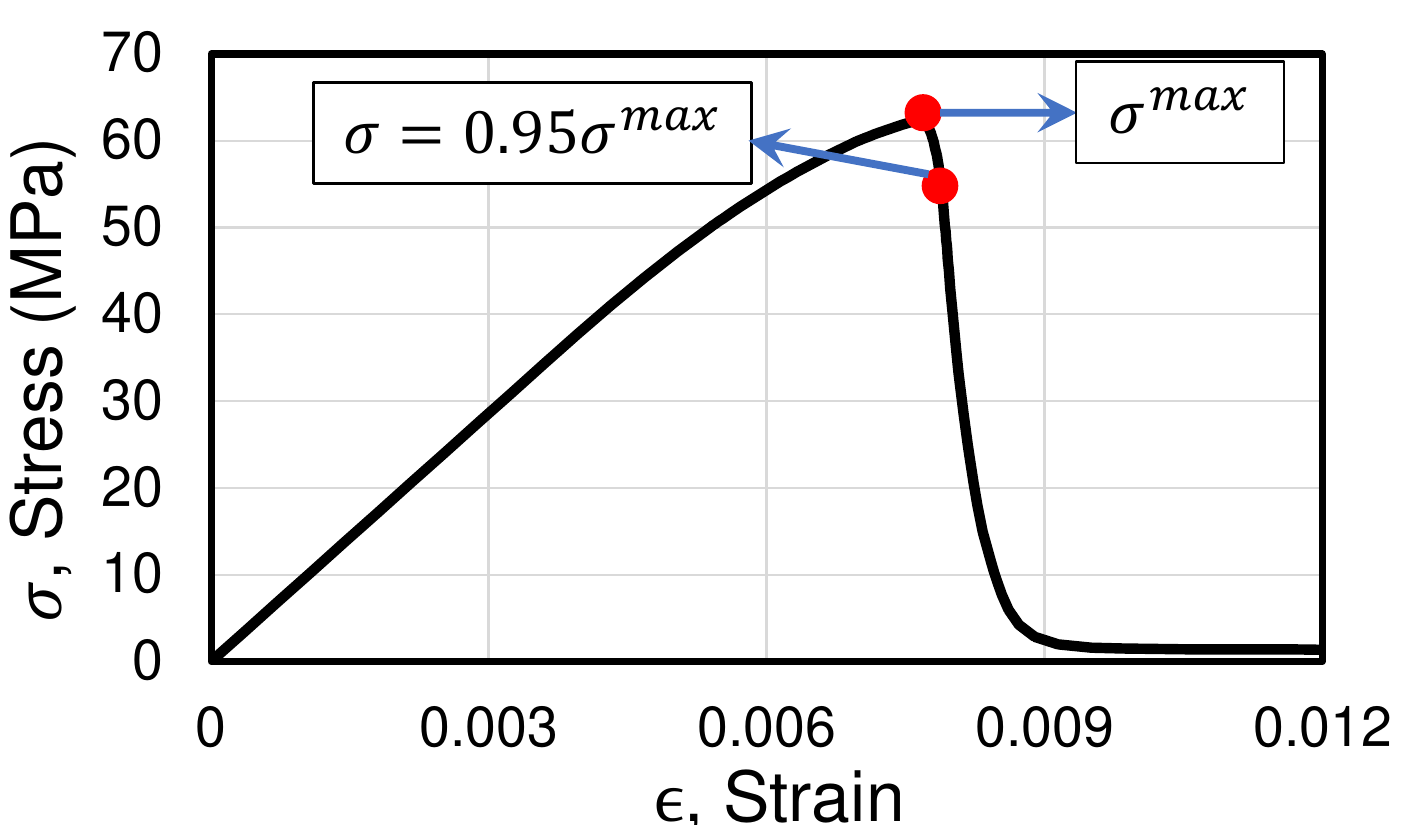}
        \caption{} \label{fig:Stress_strain}
    \end{subfigure}
    \begin{subfigure}{0.25\textwidth}
	    \centering
        \includegraphics[width=\textwidth]{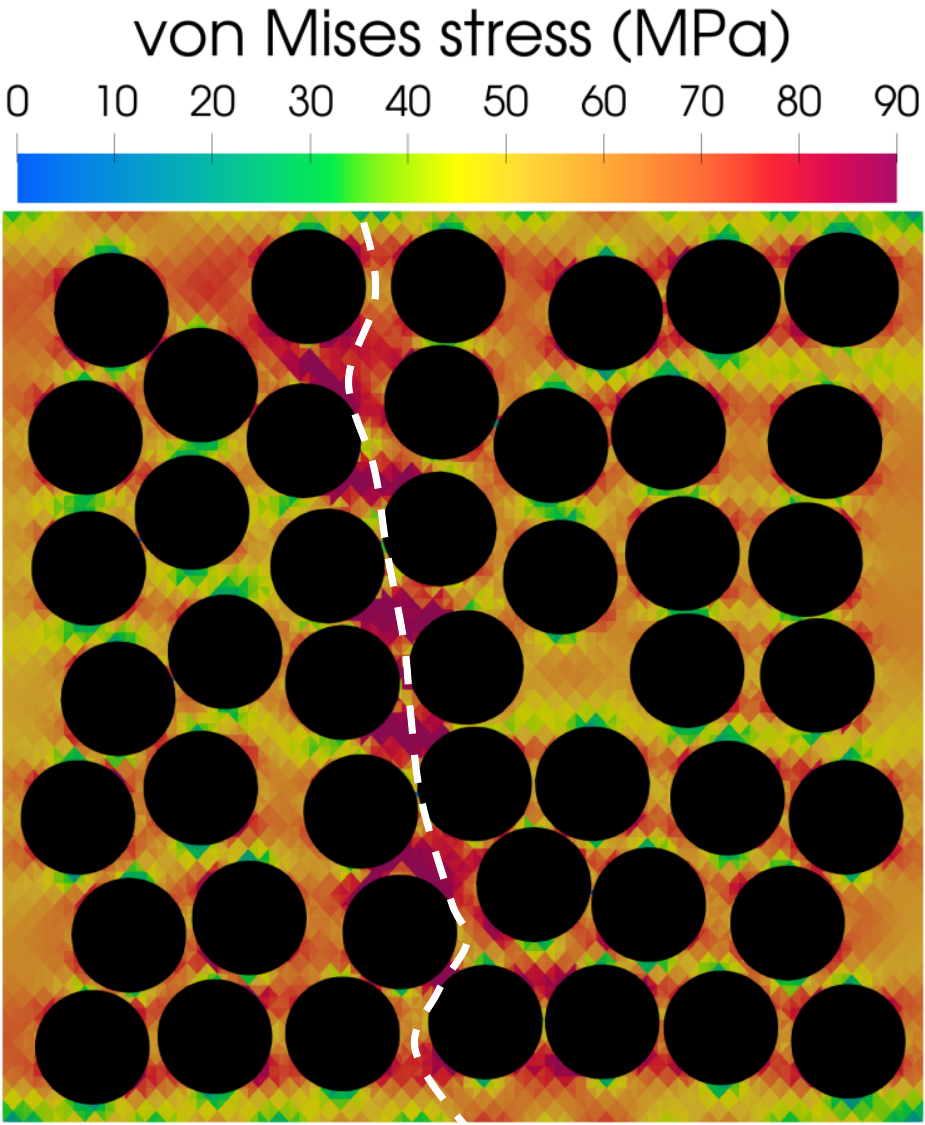}
        \caption{} \label{fig:von_Mises}
    \end{subfigure}
	\caption{(a) boundary condition imposed to RVEs in the FE simulation framework, (b) the typical stress-strain response of the RVEs, showing the ESoDI stage, and (c) von Mises stress contours at ESoDI with the white dashed line showing the crack pattern.}
	\label{FIG:mehcanical simulaiton results}
\end{figure*}

The material of interest in this work was chosen to be high-performance unidirectional carbon fiber-reinforced polymer (CFRP) composites. Several RVEs of this composites were generated randomly, as shown in Figure \ref{fig:boundary_condition}, based on a developed algorithm described in Section \ref{sectoin:RVE}. The mechanical responses of these RVEs were first simulated using a developed FE framework to produce the images of the crack pattern based on an image of the microstructure. RVEs were then subjected to transverse tension, as illustrated in Figure \ref{fig:boundary_condition}. A typical stress-strain response of such an RVE under the applied strain is presented in \ref{fig:Stress_strain}. The developed deep learning framework in this work first translates the microstructural geometry to the von Mises stress contours at the early stage of damage initiation (ESoDI) when the load carrying capacity is decreased by 5\%, as shown in Figure \ref{fig:Stress_strain}. At that stage, the stress re-distribution pattern makes the crack path apparent, as depicted in Figure \ref{fig:von_Mises}. Then, the network translates the stress contours at ESoDI to the crack pattern. In order to achieve the highest possible accuracy, a physics-informed attention objective function was designed to guide the training. For the accuracy evaluation purpose, another attention loss function was designed and utilized. This attention loss provides a more reliable estimation of when the training should be stopped before the network overfits to the training data set. In the final stage, the true accuracy of the optimized deep learning framework is evaluated based on a validation data set, through expert supervision; the predicted crack patterns are visually inspected and the number of good, partly good, and poor predictions are counted. It is also shown that the a direct translation from microstructural geometry to crack pattern, without the presence of stress at ESoDI as the intermediate prediction, results in a poor performance.

In the following section, we briefly explain the properties of the composite of interest as well as the FE simulation framework which was used to obtain the training data. Then, a detailed explanation on a proposed algorithm used to generate several realistic RVEs are presented.

\section{High-Fidelity Simulation}
The overall behavior of high-performance CFRP composites is highly impacted by the formation of transverse micro-cracks where the applied load is perpendicular to the fiber direction. The micro-cracks result from the failure of matrix and fiber/matrix interface, forming around 5\% and 95\% of the crack path, respectively \cite{gamstedt1999micromechanisms,hobbiebrunken2006evaluation,montgomery2018multiscale}. The crack pattern greatly depends on the microstructural geometry \cite{Hernandez2020}. The microstructural geometry is characterized by high volume fraction of randomly distributed aligned and continuous fibers (i.e., typically greater than $55\%$ of the composite volume) within an epoxy matrix. The distribution of fibers is deemed to be a significant predictor of the crack pattern and can significantly impact the robustness of a numerical micro-model \cite{Hernandez2020,sepasdar2020}. Hence, a random fiber generator algorithm was developed to generate several realistic representative volume elements (RVEs) for the CFRP composites microstructure. In the RVEs, the material properties of the constituents were kept constant while the fibers' distribution varied.
The RVEs were then simulated using an FE framework to predict their crack patterns. The FE framework was designed to be fast and efficient for particulate composites simulations and provided the opportunity of simulating the many generated RVEs in the shortest possible amount of time. The FE simulation framework is explained in detail in \ref{simulation framework}.

\section{Representative Volume Elements} \label{sectoin:RVE}
The fibers' distribution in high-performance CFRP composites is compact, characterized by the adjacent fibers being close and sometimes even touch one another. The matrix cracking portion of a transverse crack occurs typically in the small spaces between the adjacent fibers, signifying the importance of the distribution of fibers' nearest neighbor distances (NNDs). Taking this distribution into account, a random fiber generator algorithm was developed to generate the microstructural representations of the CFRP composites.
The algorithm generates 2-D representations of the microstructures based on a given fiber volume fraction and a target distribution for the fibers' NNDs. In this paper, the target distribution is calculated based on an experimental image of the microstructure, captured in Prof.~Sottos lab at the University of Illinois at Urbana–Champaign \cite{montgomery2018multiscale}. The algorithm proposed here is based on iteratively applying random perturbation to randomly selected fibers, an idea initially proposed by Zacek \cite{zacek2017exploring}. However, in the algorithm proposed here, the fibers are able to significantly dislocate, the accuracy is higher, and the rate of convergence is faster.

\begin{figure*}
	\centering
	\includegraphics[width=0.8\textwidth]{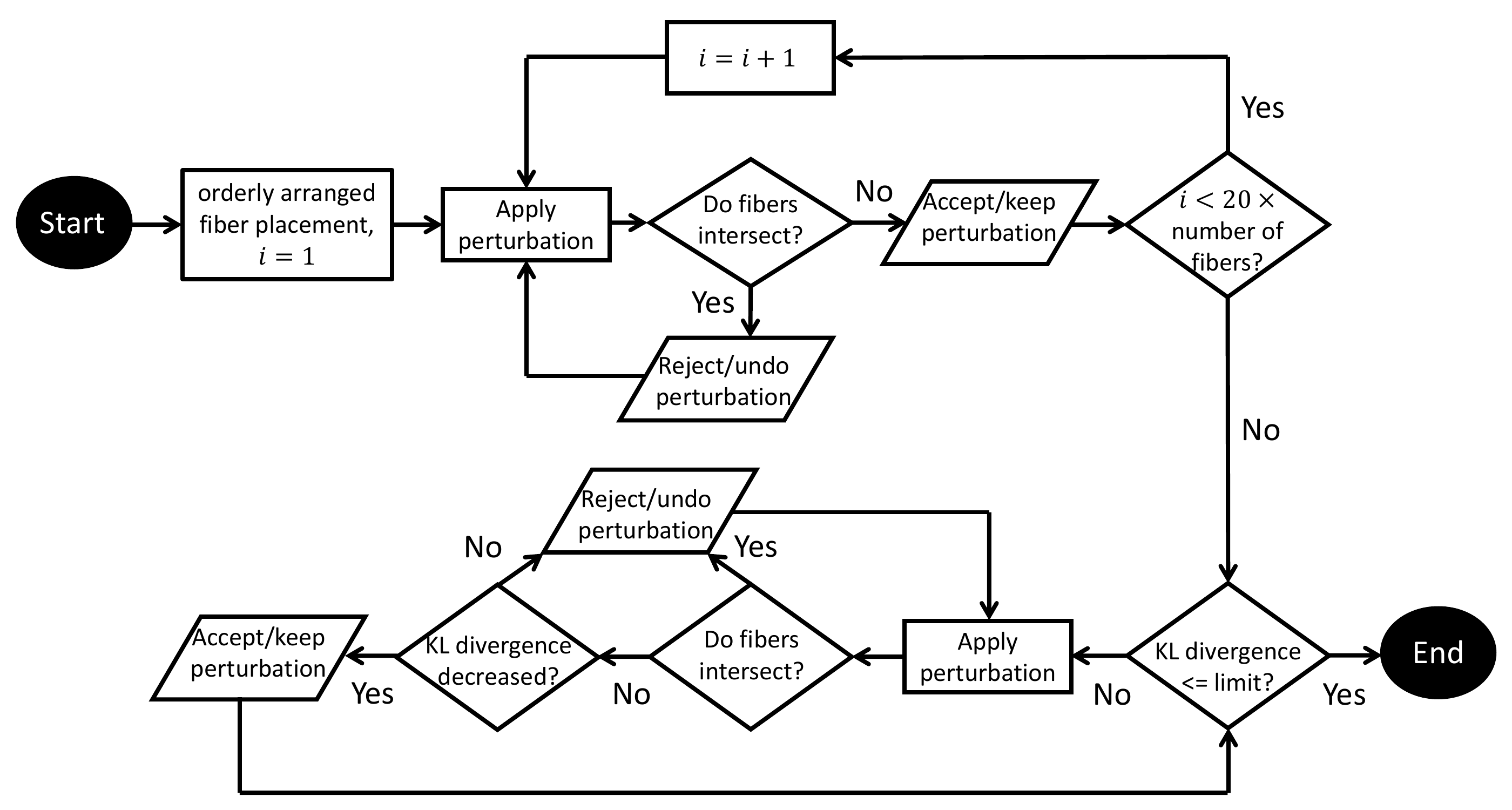}
	\caption{The flowchart of the proposed algorithm for the random RVE generation based on a target distribution for fibers' NND. In the algorithm, the similarity between the generated and target fibers' NND distributions is checked using KL divergence.}
	\label{FIG:RVE_algorithm}
\end{figure*}

In this algorithm, the fibers are initially orderly arranged in a staggered pattern within a specified domain. Then, a fiber is randomly selected and randomly perturbed within a specified range. If the fiber does not intersect with other fibers, the perturbation is accepted. This process is repeated several times until the number of accepted perturbations exceeds 20 times the number of fibers within the RVE. At this stage, the initially uniformly distributed fibers inside the RVE are perfectly shuffled. Then, a second round of random fiber selection and perturbation is applied. This time, however, an additional condition is imposed to the acceptance criterion for the perturbation; aside from the intersection condition, the perturbation is also required to improve the distribution of the fibers NND to be accepted. The similarity between the distributions is checked using the Kullback–Leibler (KL) divergence measure. The second round of perturbation is repeated until the resulting distribution for the fibers NND matches with the target distribution. The flowchart of the proposed algorithm is illustrated in Figure \ref{FIG:RVE_algorithm}. In the proposed approach, a distance matrix for the fibers is calculated at each iteration to facilitate checking fibers intersection condition and measuring the fibers NND.

\begin{figure*}
    \centering
	\begin{subfigure}{0.25\textwidth}
	    \centering
        \includegraphics[width=\textwidth]{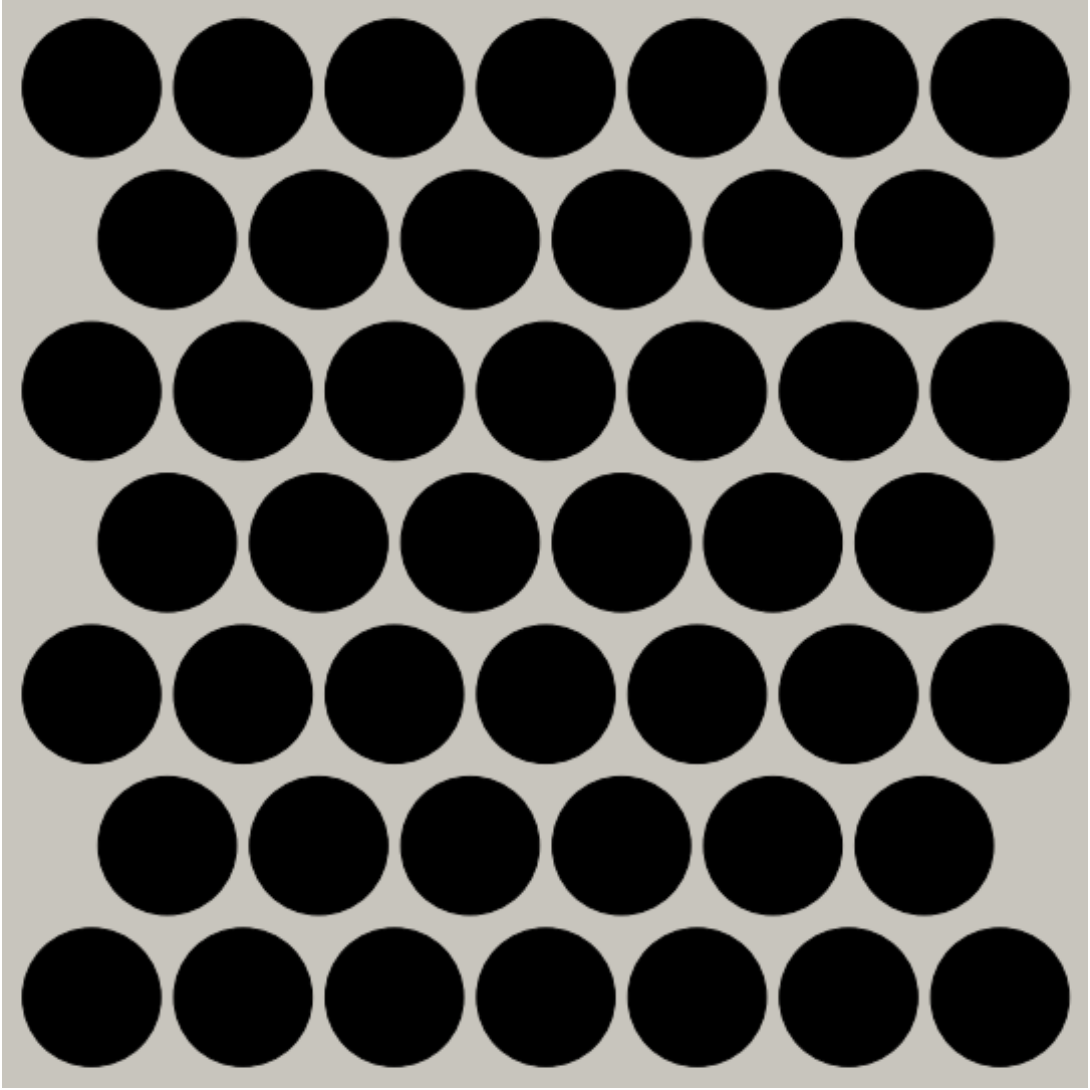}
        \caption{} \label{fig:RVE_result-a}
    \end{subfigure}
    \begin{subfigure}{0.4\textwidth}
	    \centering
        \includegraphics[width=\textwidth]{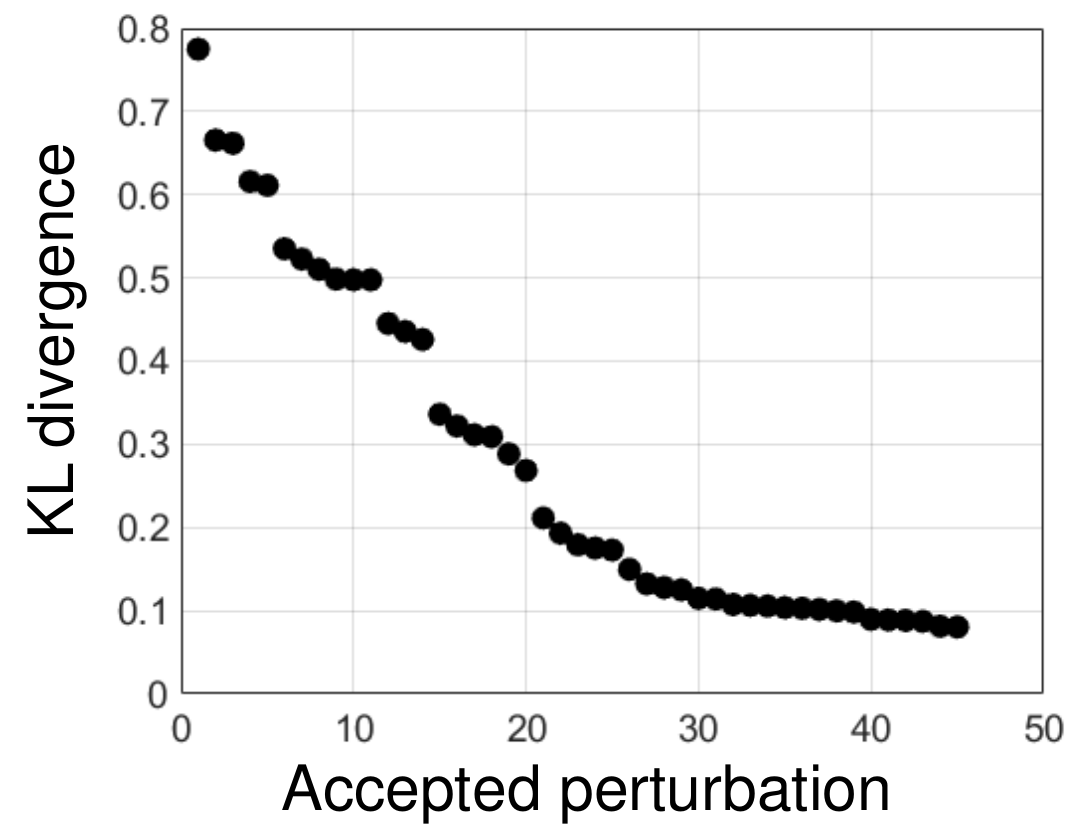}
        \caption{} \label{fig:RVE_result-b}
    \end{subfigure}
    \\
    \begin{subfigure}{0.25\textwidth}
	    \centering
        \includegraphics[width=\textwidth]{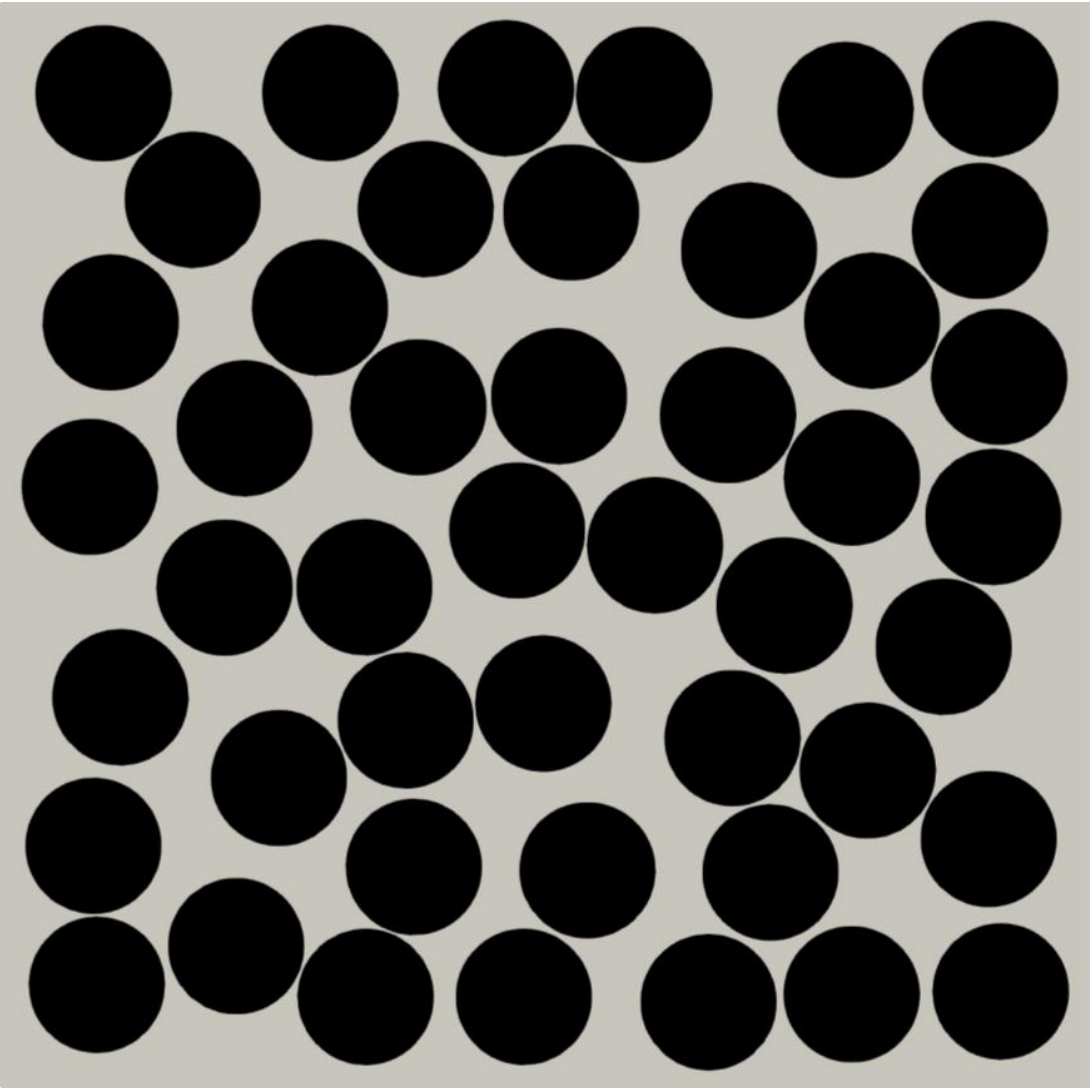}
        \caption{} \label{fig:RVE_result-c}
    \end{subfigure}
    \begin{subfigure}{0.4\textwidth}
	    \centering
        \includegraphics[width=\textwidth]{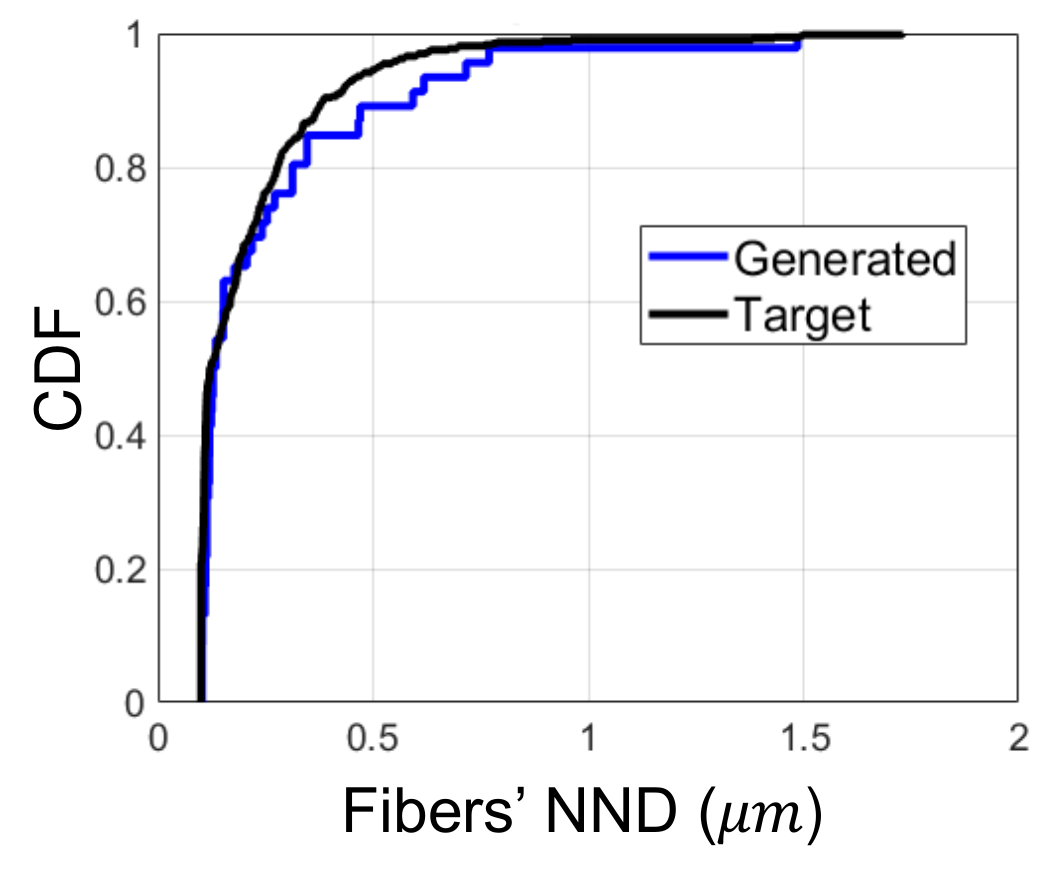}
        \caption{} \label{fig:RVE_result-d}
    \end{subfigure}
	\caption{Illustration of the procedure for generating an RVE: (a) initial orderly arranged fibers placement, (b) decrease of KL divergence for the generated and target fibers' NND distributions with the increase of the number of accepted perturbations, (c) the final generated RVE, and (d) the comparison between the fibers' NND distribution of the final generated RVE and the target distribution based on their cumulative density functions (CDFs).}
	\label{FIG:RVE generator}
\end{figure*}

The described algorithm was used to generate 4500 square RVEs with dimensions $54\mu m \times 54\mu m$. The RVEs contained 46 fibers with a diameter of $7\mu m$ and a volume fraction of 57.5\%. The initial regular distribution of fibers at the start of algorithm is shown in Figure \ref{fig:RVE_result-a}. After the application of the algorithm, the distribution of fibers NND is improved as the number of accepted perturbation increases, as shown in Figure \ref{fig:RVE_result-b}. An example of a generated RVE and the similarity of its fibers' NND distribution with the target distribution are illustrated in Figure \ref{fig:RVE_result-c} and Figure \ref{fig:RVE_result-d}, respectively. The generated RVEs were then simulated using the developed FE framework under transverse tension.

\section{Deep Learning Framework}
\begin{figure*}
	\centering
	\includegraphics[width=0.8\textwidth]{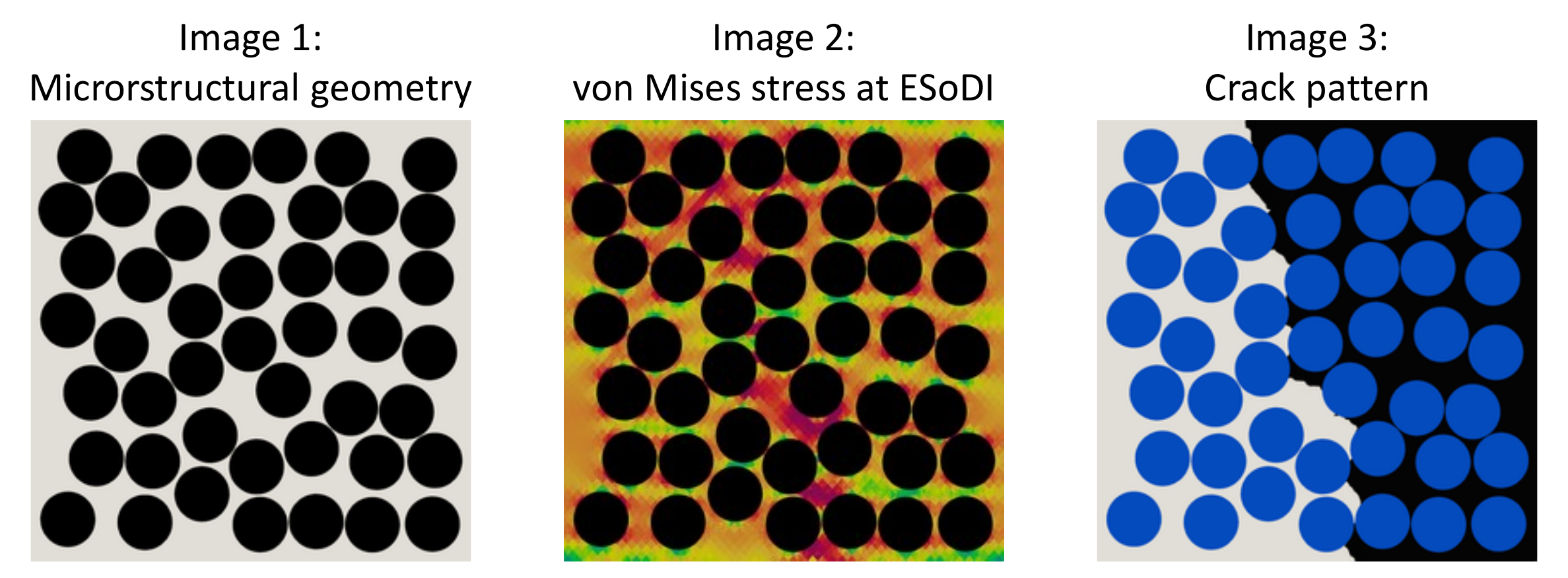}
	\caption{The triple images required for the training of the proposed deep learning framework predict the crack pattern image based on an image of the microstructure.}
	\label{FIG:Triple_images}
\end{figure*}

The results of the 4500 FE simulated RVEs were used as a data set to train a deep learning framework to generate the crack pattern based on an image of the microstructure. As described earlier, the framework also generates the matrix von Mises stress distribution at ESoDI as an intermediate prediction. It is reminded that ESoDI is defined as the post-failure state where the load carrying capacity is decreased by 5\%, computed based on the RVE stress-strain response as shown in Figure \ref{fig:Stress_strain}. The intermediate prediction is necessary for the framework to achieve a high accuracy in the ultimate crack pattern prediction. Hence, the network is trained based on the triple images of microstructure (Image 1), von Mises stress at ESoDI (Image 2), and crack pattern (Image 3) as illustrated in Figure \ref{FIG:Triple_images}.

There are three reasons behind using the image format instead of the real-valued field outputs (e.g., the actual von Mises stress values):
\begin{enumerate}
    \item The input to the network has to be in the form of regularly dispersed 2-D grids, represented by a 2-D tensor, so that the convolutional operation can be utilized. The field outputs of the FE simulations, however, correspond to the nodal values of the FE mesh, the locations of which are not regularly distributed in the space. The complexity of geometry prevents applying an FE mesh with a perfectly uniform distribution of nodes. For more information on the FE mesh, the reader is referred to \ref{simulation framework}.
    \item Converting the irregularly distributed nodal field values of the FE mesh into uniformly distributed grids requires complex polynomial fitting techniques. Loss of information, especially in the zones with high-stress gradients (i.e., located between the neighbouring fibers), is inevitable, inducing some degrees of inaccuracy. It was observed that the resulting inaccuracy was less significant when the contour plot of a continuous field output was saved in the image format.
    \item The process of getting field output contour images from the commercial FE simulation programs is quick and straight forward. 
\end{enumerate}

\subsection{Data Pre-processing}
The utilized FE simulation framework stores the results in the VTK format, which contains both the geometry of the FE mesh and the simulation results as the nodal values. The VTK results were plotted using ParaView, an open-source multi-platform application for scientific visualization. The material tag, von Mises stress, and displacement contours were used to plot the microstructural geometry (Image 1), matrix von Mises stress distribution at ESoDI (Image 2), and crack pattern (Image 3), respectively, in the format of high-quality JPEG images. 

The microstructures' images were saved such that the matrix and fibers were differentiated through a specified distinct coloration of white and black. In the images of crack patterns, the fibers are colored blue, while the right and left sides of the cracks were colored white and black, respectively. The dimensions of the images were then reduces to 256$\times$256 pixels. 
The specified dimensions are consistent with the mesh size used in the FE simulation framework and provide enough accuracy for presenting the simulation results.

The images were then split into mutually exclusive training and validation data sets, containing 4000 and 500 triple image samples, respectively. As a data augmentation technique, the number of samples in the training data set was doubled by flipping all the triple images vertically. Aside from increasing the size of the training data set at no cost, this data augmentation also encourages the networks to become robust to symmetry and learn features from the overall geometrical pattern in addition to the morphology of local features. 

\subsection{Networks Architecture}
\begin{figure*}
	\centering
	\includegraphics[width=\textwidth]{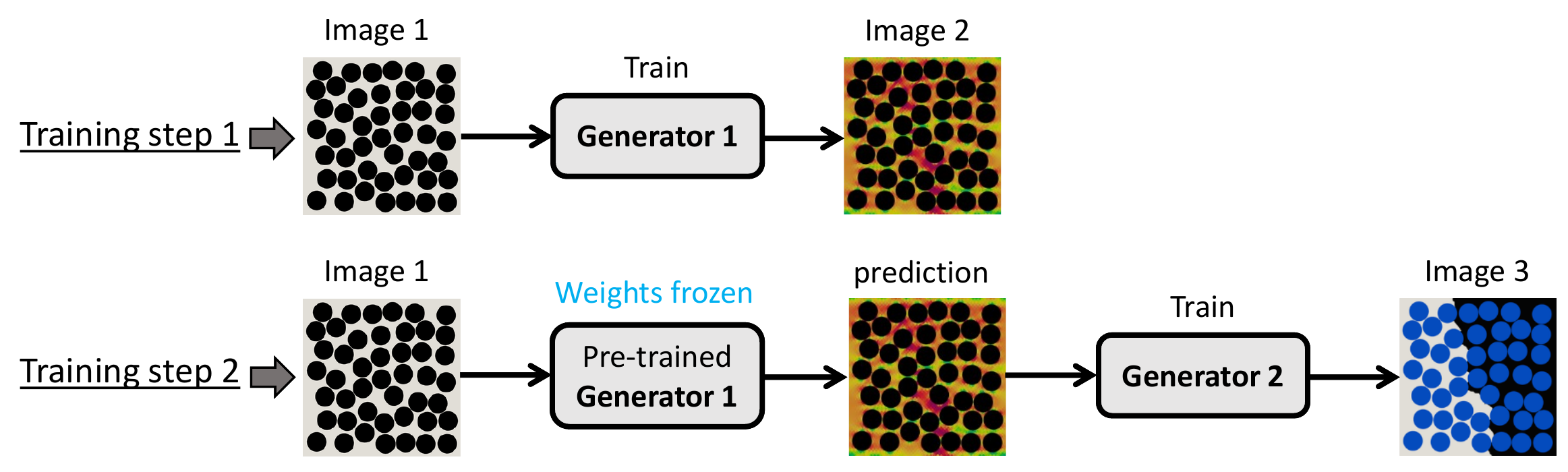}
	\caption{The sequential training approach utilized in the proposed deep learning framework for training Generator 1 and Generator 2.}
	\label{FIG:Training_strategy}
\end{figure*}

The developed deep learning framework includes two stacked generator networks that are sequentially trained in a supervised fashion:
\begin{enumerate}
    \item Generator 1 translates the image of microstructure to the image of the matrix von Mises stress distribution at ESoDI.
    \item Generator 2 translates the image of the von Mises stress contours at ESoDI to the crack pattern image. This generator is trained using the outputs of Generator 1 as the input.
\end{enumerate}
The described sequential training of Generator 1 and Generator 2 is schematically illustrated in Figure \ref{FIG:Training_strategy}.

\begin{figure*}
	\centering
	\includegraphics[width=\textwidth]{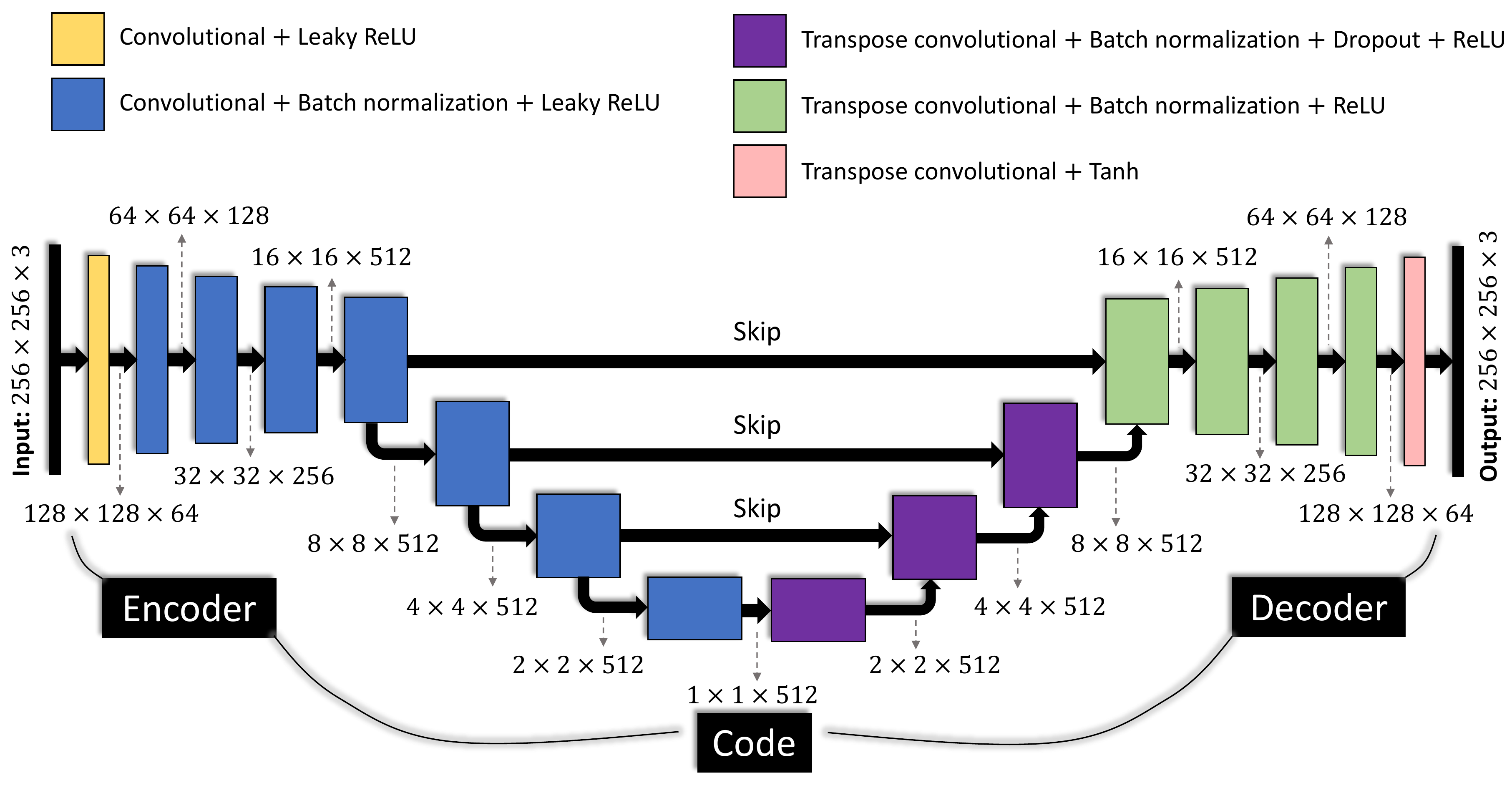}
	\caption{The architecture of the modified U-Net network used for both Generator 1 and Generator 2.}
	\label{FIG:Network}
\end{figure*}

A modified U-Net architecture was used for both Generator 1 and Generator 2. U-Net is a fully convolutional neural network autoencoder that was first proposed by Ronneberger et al. \cite{ronneberger2015u} in 2015. This architecture is proven to be optimal for image translation applications \cite{zhou2018unet++,ibtehaz2020multiresunet,koeppe2020intelligent,tang2021deep}. The network is composed of two branches of convolutional layers, namely Encoder and Decoder, as illustrated in Figure \ref{FIG:Network}. Encoder transforms a high-dimensional $256\times 256 \times 3$ pixel-image into a lower-dimensional $1\times 1 \times 512$ vector of features that best describes the input image. This lower-dimensional vector is called Code and contains encrypted information and features of the input image that are the significant predictors of the output image. The Decoder network is responsible for translating the encrypted information of the Code vector into the desired output image.

Each block of Encoder contains a convolutional layer with a kernel size of 4 and a stride of 2. The convolutional layer is then followed by a batch normalization (except for the first block), the output of which is passed through the Leaky ReLU activation function. The blocks output dimensions, including the number of applied convolutional filters as the third dimension, are shown in Figure \ref{FIG:Network}.

Each block of Decoder contains a transpose convolutional layer, with the same properties and number of filters as the convolutional layer of the corresponding block in Encoder. Batch normalization followed by a ReLU activation are then applied. A drop-out of 0.5 is also considered for the first three blocks of Decoder to further boost the robustness of the predictions. The final layer of the U-Net architecture is a transpose convolutional operation with a Tanh activation that is responsible to transform the $128\times 128\times 64$ output of the previous block into a $256\times 256\times 3$ image, the final output of the network.

In the original U-Net architecture, there are skip connections between all the blocks of Encoder and the corresponding block of Decoder to facilitate learning and the generation of complex/realistic images. In this work, however, the skip connections are applied only to the first three inner blocks of Encoder and Decoder to prevent overfitting. 

\subsection{Networks Training Objective Functions} \label{section: training loss}
For Generator 2, mean absolute error (MAE) loss, $\mathcal{L}_{MAE}$, is considered as the objective function. The MAE is a typical and simple loss function used in image-to-image translation problems. The simple MAE loss can provide high accuracy for Generator 2 since predicting the crack pattern based on the von Mises stress at ESoDI is a simple task. 

Generator 1, on the other hand, carries the heaviest burden in the process of crack pattern prediction. Generator 1 has to learn several complicated and non-obvious local and global patterns in the microstructural geometry, figure out which patterns are the significant predictors of crack pattern, and translate the patterns to the complex and nontrivial von Mises stress distribution at ESoDI.
Hence, for Generator 1, two different objective functions are considered, and their performances are compared. The considered objective functions were: 
\begin{enumerate}
    \item MAE loss, which is evaluated using the predicted and target images.
    \item Attention loss, which is a modified MAE that magnifies the absolute error for the pixels that are more important in determining the crack pattern.
\end{enumerate}

\begin{figure*}
    \centering
	\begin{subfigure}{0.3\textwidth}
	    \centering
        \includegraphics[width=\textwidth]{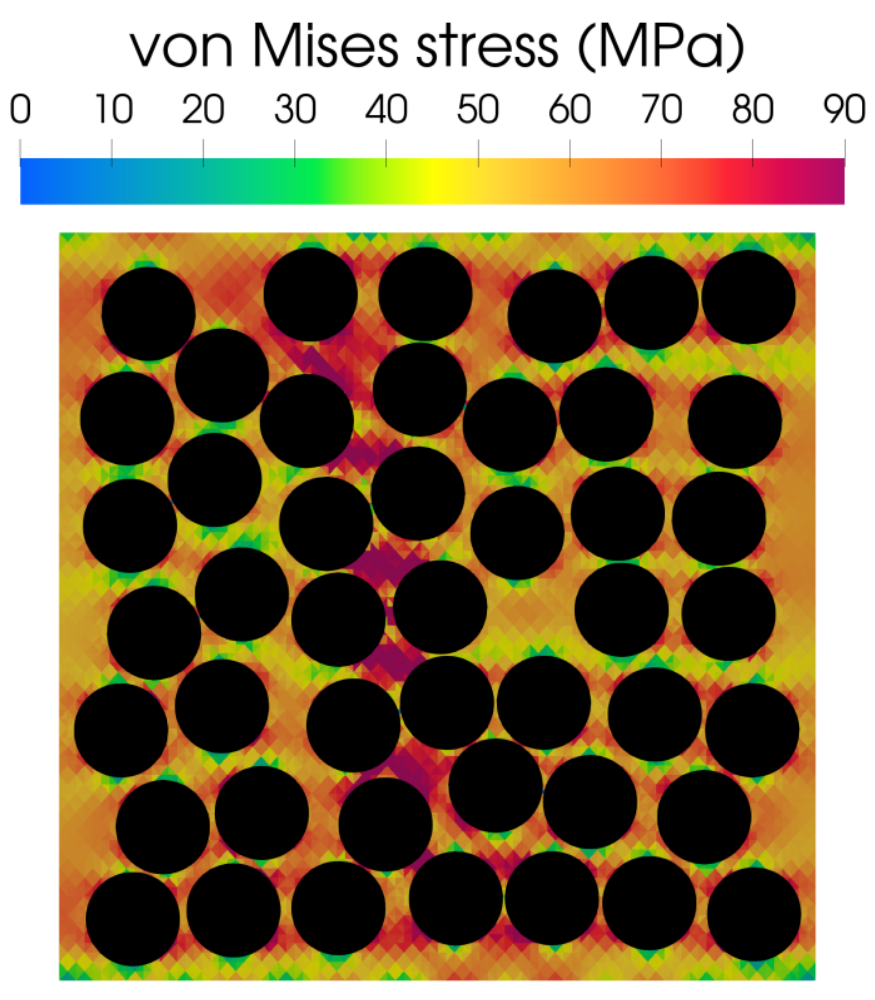}
        \caption{} \label{fig:large_stress_von}
    \end{subfigure}
	\begin{subfigure}{0.3\textwidth}
	    \centering
        \includegraphics[width=\textwidth]{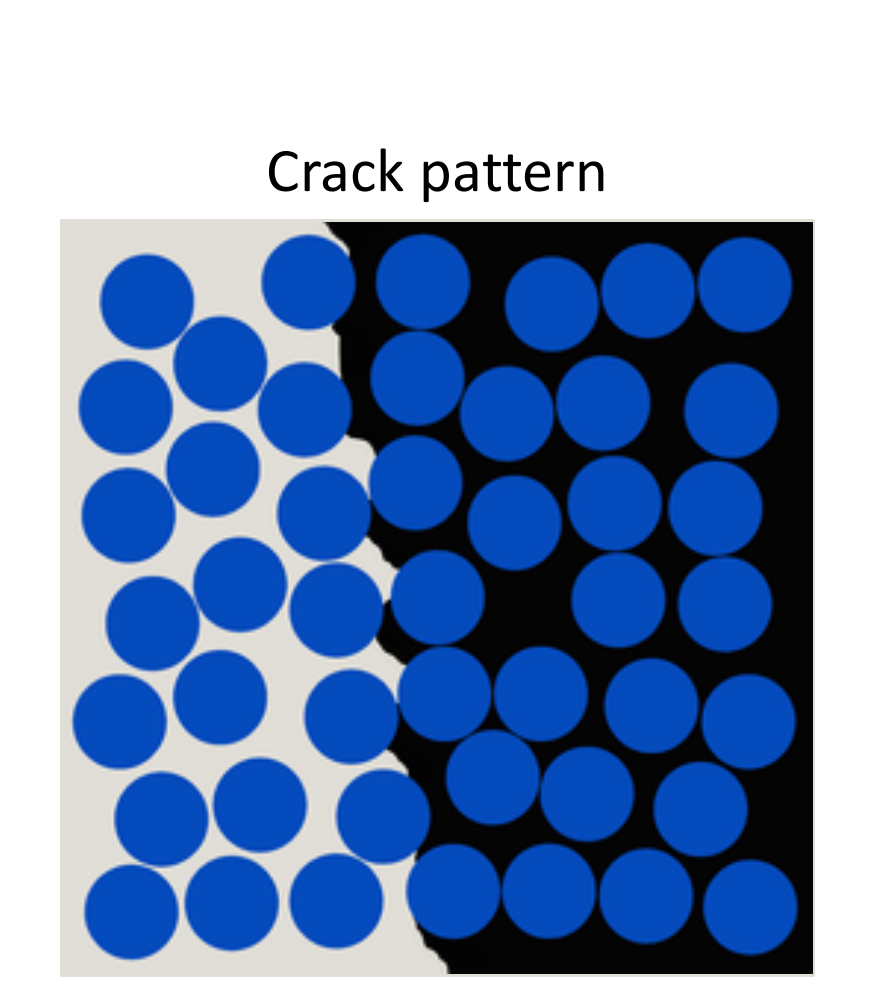}
        \caption{} \label{fig:large_stress_crack}
    \end{subfigure}
	\begin{subfigure}{0.3\textwidth}
	    \centering
        \includegraphics[width=\textwidth]{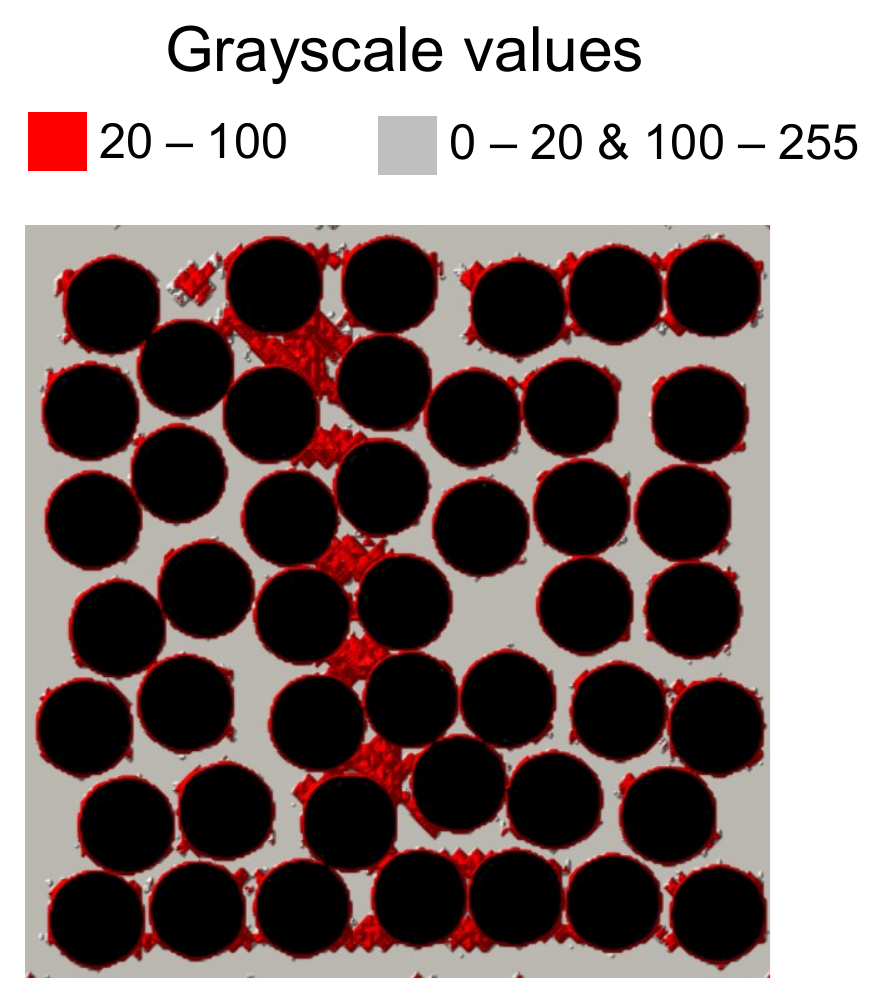}
        \caption{} \label{fig:large_stress_grey}
    \end{subfigure}
    \\
    \begin{subfigure}{0.45\textwidth}
	    \centering
        \includegraphics[width=\textwidth]{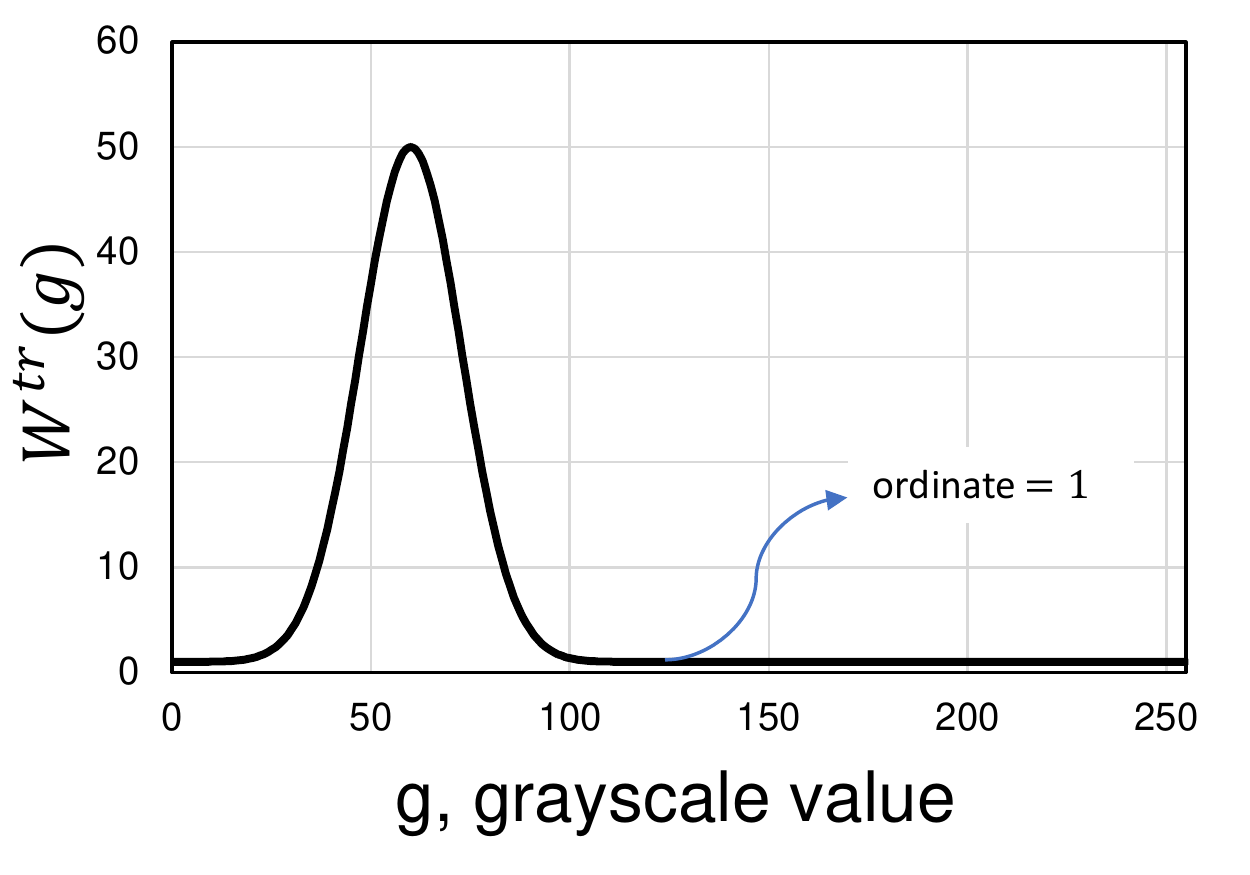}
        \caption{} \label{fig:attention_loss}
    \end{subfigure}
    \begin{subfigure}{0.45\textwidth}
	    \centering
        \includegraphics[width=\textwidth]{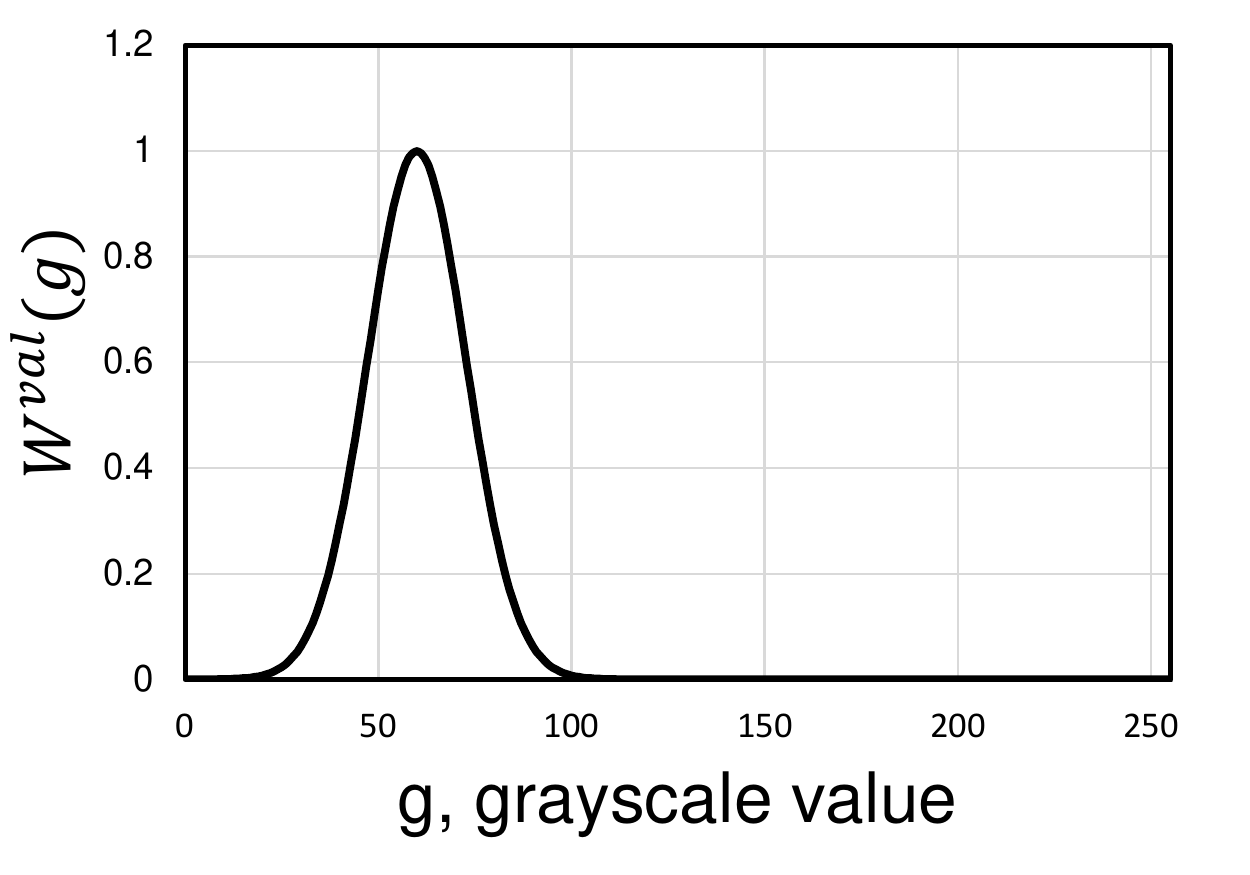}
        \caption{} \label{fig:attention_accuracy}
    \end{subfigure}
	\caption{(a), (b) An illustration of high values of the von Mises stress distribution at ESoDI, being the significant predictors of the crack pattern, and (c) the range of grayscale values, corresponding to the high-stress zones that can determine the crack pattern.
    (b) and (c) the weight functions incorporated in the attention losses, used for Generator 1's training and accuracy evaluation, respectively.}
	\label{FIG:losses}
\end{figure*}

As it was described in Section \ref{section: overview}, the stress distribution at ESoDI facilitates the identification of the crack path, as illustrated in Figure \ref{fig:large_stress_von} and Figure \ref{fig:large_stress_crack}. Comparing these two figures demonstrates that the pixel values, corresponding to large magnitudes of stress are the significant predictors of crack pattern. These high-stress pixels have grayscale values between 20 and 100 and can alone determine the crack pattern, as depicted in Figure \ref{fig:large_stress_grey}. In this paper, the grayscale value of a pixel is calculated by averaging the pixel's red, green, and blue (RGB) channels' values.

Therefore, the attention loss is designed to penalize MAE by increasing the contribution of the absolute errors, corresponding to the zones with high-stress concentration. More specifically, the attention loss is a weighted MAE where weights greater than 1 are considered for the pixels, representing large values of von Mises stress when the mean is calculated. This modification to the MAE loss increases the gradient of loss for high-stress pixels and encourages the network to pay more attention to the specific areas of the image that determine the crack pattern. The other areas of the ESoDI stress images, comprising the majority of the pixels, are not significant predictors of the crack pattern. Hence, the attention loss prevents the majority of the network's intelligence and effort to be spent in the accurate prediction of less important areas.


To design the attention training loss to function as described above, the following Gaussian function is used to calculate a weight, $W^{tr}$, based on a given grayscale pixel value, $g$:
\begin{equation}\label{eq:training loss}
    W^{tr}(g) = \alpha e^{-\frac{(g-\beta)^2}{2\gamma^2}}+1
\end{equation}
where $g$ is the pixel's grayscale value, 
$\alpha+1$ and $\beta$ specify the ordinate and abscissa of the function's turning point, respectively, whereas $\gamma$ controls the sharpness of the turning point. The constant $\beta$ was set equal to the pixel value for the largest magnitude of stress, whereas $\gamma$ was selected such that the range of interest for the pixel values (i.e., from 20 to 100) are lied underneath the nonlinear portion of the function to be assigned weights greater than 1. $\alpha$ controls the significance of the pixel values that correspond to large stress magnitudes and is a tunable hyper parameter of the network. The calibrated values for the parameters of the weight function presented by Equation \ref{eq:training loss} are $\alpha=50$, $\beta=60$, and $\gamma=0.1$, resulting in a weight function as illustrated in Figure \ref{fig:attention_loss}. Subsequently, the attention training objective function is calculated as
\begin{equation}\label{eq:attention MAE}
    \mathcal{L}^{tr}_{att}=Mean\left(W^{tr}(\mathbf{Y})\circ |\hat{\mathbf{Y}}-\mathbf{Y}|\right)
\end{equation}
where $\mathbf{Y}$ and $\hat{\mathbf{Y}}$ are the target and generated images, respectively, and the symbol $\circ$ is the Hadamard product operator.

\subsection{Validation}
The accuracy is evaluated based on the validation data set to reflect the true performance of the networks. In the proposed framework, the accuracy needs to be evaluated twice;
when Generator 1 is trained, and when Generator 2 is trained using the outputs of pre-trained Generator 1.

\subsubsection{Generator 1 accuracy}\label{gen 1 acc}
As stated earlier, the accuracy of the deep learning framework in predicting the crack pattern greatly depends on the performance of Generator 1, producing von Mises stress distribution. Generator 1, however, is susceptible to overfitting, and hence, its training must be stopped at an appropriate training epoch to prevent over-training. As a result, an appropriate measure is required for evaluating the true accuracy of Generator 1 after the application of each epoch.

As described earlier in Section \ref{section: training loss}, the accuracy of the deep learning framework depends upon the distribution pattern of only a few pixels in Generator 1 output, representing large values of von Mises stress (Figure \ref{fig:large_stress_von} and Figure \ref{fig:large_stress_crack}). Therefore, The optimal epoch for Generator 1 cannot be estimated through a pixel-wise comparison (e.g., $\mathcal{L}_{MAE}$) between generated and target images.
The MAE loss treats all pixels the same way, without any discrimination, allowing for the contribution of non-important regions. Therefore, an attention MAE loss for validation ($\mathcal{L}^{val}_{att}$), similar to that used in the training of Generator 1, is proposed. $\mathcal{L}^{val}_{att}$ only targets the pixels that correspond to large values of stress through a weight function $W^{val}$ defined as
\begin{equation}\label{eq:validation loss}
    W^{val}(g) = e^{-\frac{(g-\beta)^2}{2\gamma^2}}
\end{equation}
$W^{val}$ has a value in the range between 0 and 1, computed based on the importance of a given pixel, as illustrated in Figure \ref{fig:attention_accuracy}. A weighted MAE evaluated using the $W^{val}$ function only includes the contribution of important zones and pixels. The weighted MAE omits the contribution of the portions of the image, which do not contribute to crack pattern identification. The attention loss for validation, $\mathcal{L}^{val}_{att}$, can then be calculated using Equation \ref{eq:attention MAE} with $W^{tr}$ replaced by $W^{val}$. The evolution of $\mathcal{L}^{val}_{att}$ during the training is then used to find the epoch that leads to optimized Generator 1. After the optimized Generator 1 is determined, its outputs are used to train Generator 2.

\subsubsection{Generator 2 accuracy} \label{gen 2 acc}
It is reminded that the task of Generator 2 is quite simple and this network can easily achieve a high accuracy with the application of only a few epochs. In other words, if Generator 1 predicts accurate stress distribution at ESoDI, the correctness of the predicted crack pattern by Generator 2 is guaranteed. Generator 2 does not have the overfitting issue and provides consistent predictions based on both validation and training data sets during the training. Hence, its accuracy is evaluated after the application of enough training epochs, when the average MAE loss of the validation data set does not decrease with the application of more epochs. 

\begin{figure*}
	\centering
	\includegraphics[width=0.5\textwidth]{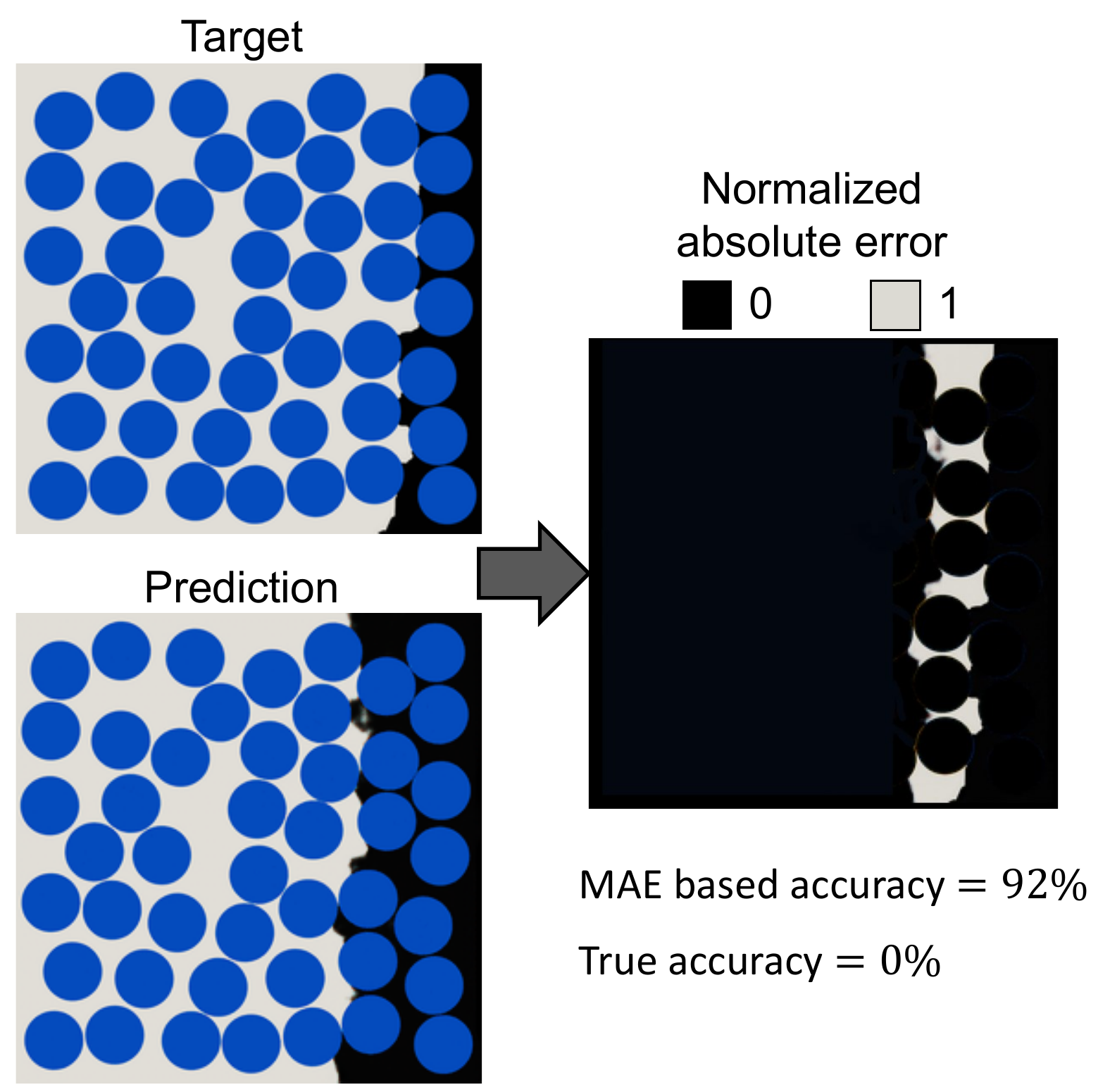}
	\caption{An example to illustrate why $\mathcal{L}_{MAE}$ cannot be used for the accuracy evaluation of Generator 2. MAE estimates a high accuracy for the predicted image while the true accuracy is zero.}
	\label{FIG:MAE_issue}
\end{figure*}

\begin{figure*}
	\centering
	\includegraphics[width=0.4\textwidth]{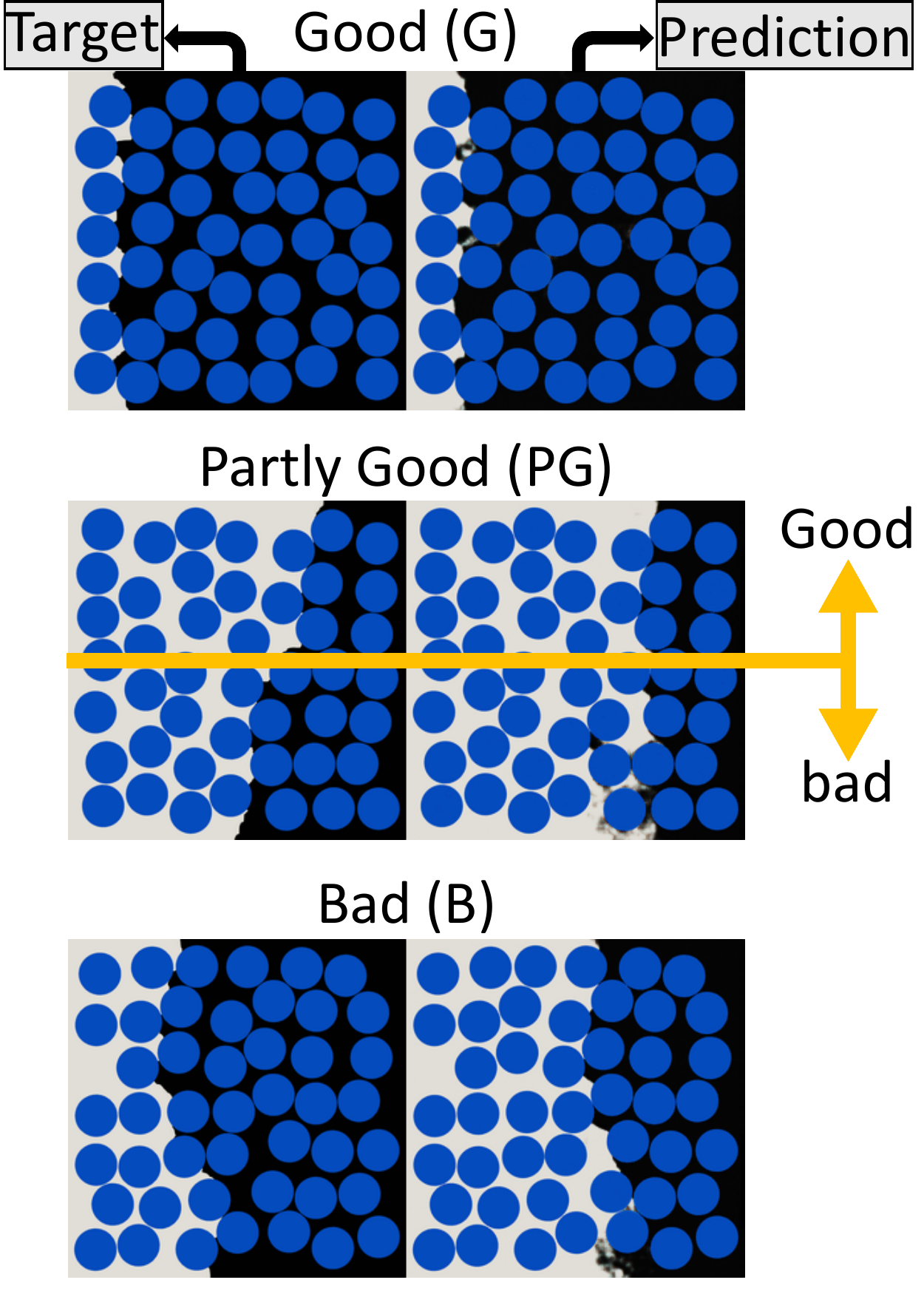}
	\caption{Illustration of good (G), partly good (PG), and bad (B) crack pattern predictions by the deep learning framework.}
	\label{FIG:good-bad-ugly}
\end{figure*}

The true accuracy of Generator 2 (i.e., also represents the accuracy of the proposed deep learning framework) should reflect the correctness of the predicted crack pattern. Therefore, pixel-wise comparison approaches such as MAE can be misleading. The reason is that a wrong predicted crack path can still result in a small $\mathcal{L}_{MAE}$, suggesting a high accuracy, while the true accuracy is zero. This point is schematically illustrated in Figure \ref{FIG:MAE_issue}.
The only accurate way of calculating the true accuracy is through expert supervision, by visual inspect the predicted and target crack patterns using the entire validation data set samples. In the visual inspection, the predictions were separated into three categories of ``good" (G), ``partly good" (PG), and ``bad" (B) predictions, as illustrated in Figure \ref{FIG:good-bad-ugly}. In the PG category, around half-length of the predicted crack pattern is ``good" while the rest is ``bad". Based on the number of samples in each category, $n$, the true accuracy, $A\%$, was calculated by
\begin{equation}
    A\%= \frac{n^{G} + 0.5\times n^{PG}}{n^{G} + n^{PG} + n^{B}}\times 100
\end{equation}
In the above equation, partial credit is given to the PG predictions through a multiplier of 0.5.

\section{Results and Discussion} \label{sectoin:results}
In this section, the training results and performance of the proposed deep learning framework are discussed. The training of both generators was conducted by using the Adam optimizer with the application of 120 epochs.  The training of the networks were conducted on the Cascades cluster at Advanced Research Computing at Virginia Tech, using an NVIDIA V100 GPU. In the following subsections, first, the training of Generator 1 is explained and the results of training with the MAE and attention losses are compared. Then, the training results of Generator 2 is presented and the ultimate accuracy of the proposed framework for both cases of Generator 1 training scenarios (i.e., trained with the MAE and attention losses), is evaluated. Finally, the framework's performance is compared versus a simplified model where a single-generator is responsible to directly translate from microstructural geometry to crack pattern.

\subsection{Generator 1 Training Result}
The first step in the sequential training of the proposed deep learning framework was training Generator 1. It is reminded that Generator 1 is trained using the images of microstructure as input and the corresponding images of von Mises stress distribution at ESoDI as output. Two different objective functions, namely, MAE and attention losses were considered. The accuracy of both cases were evaluated using the attention loss, $\mathcal{L}^{val}_{att}$, as presented in Section \ref{gen 1 acc}. 

\begin{figure*}
    \centering
    \begin{subfigure}{0.8\textwidth}
	    \centering
        \includegraphics[width=\textwidth]{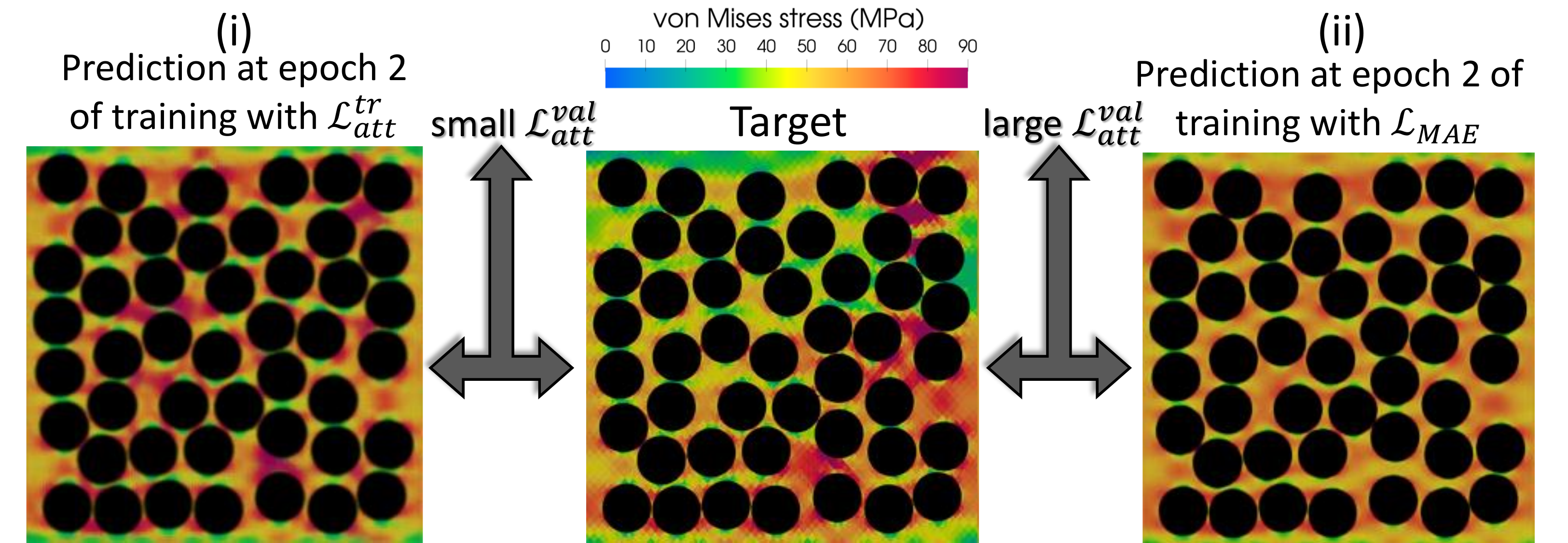}
        \caption{} \label{fig:initial_epoch_result}
    \end{subfigure}
    \\
	\begin{subfigure}{0.45\textwidth}
	    \centering
        \includegraphics[width=\textwidth]{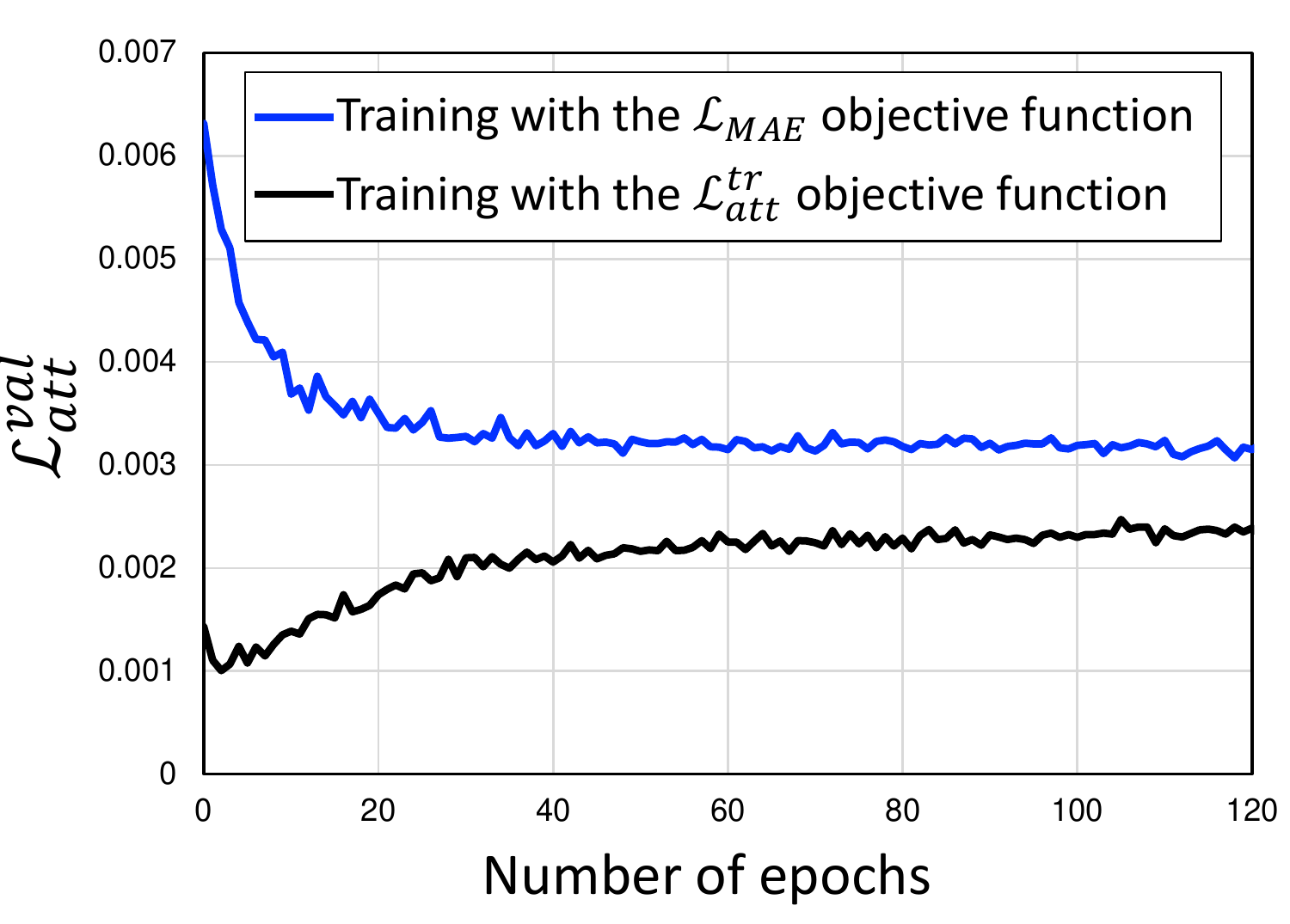}
        \caption{} \label{fig:eval_loss}
    \end{subfigure}
    \begin{subfigure}{0.45\textwidth}
	    \centering
        \includegraphics[width=\textwidth]{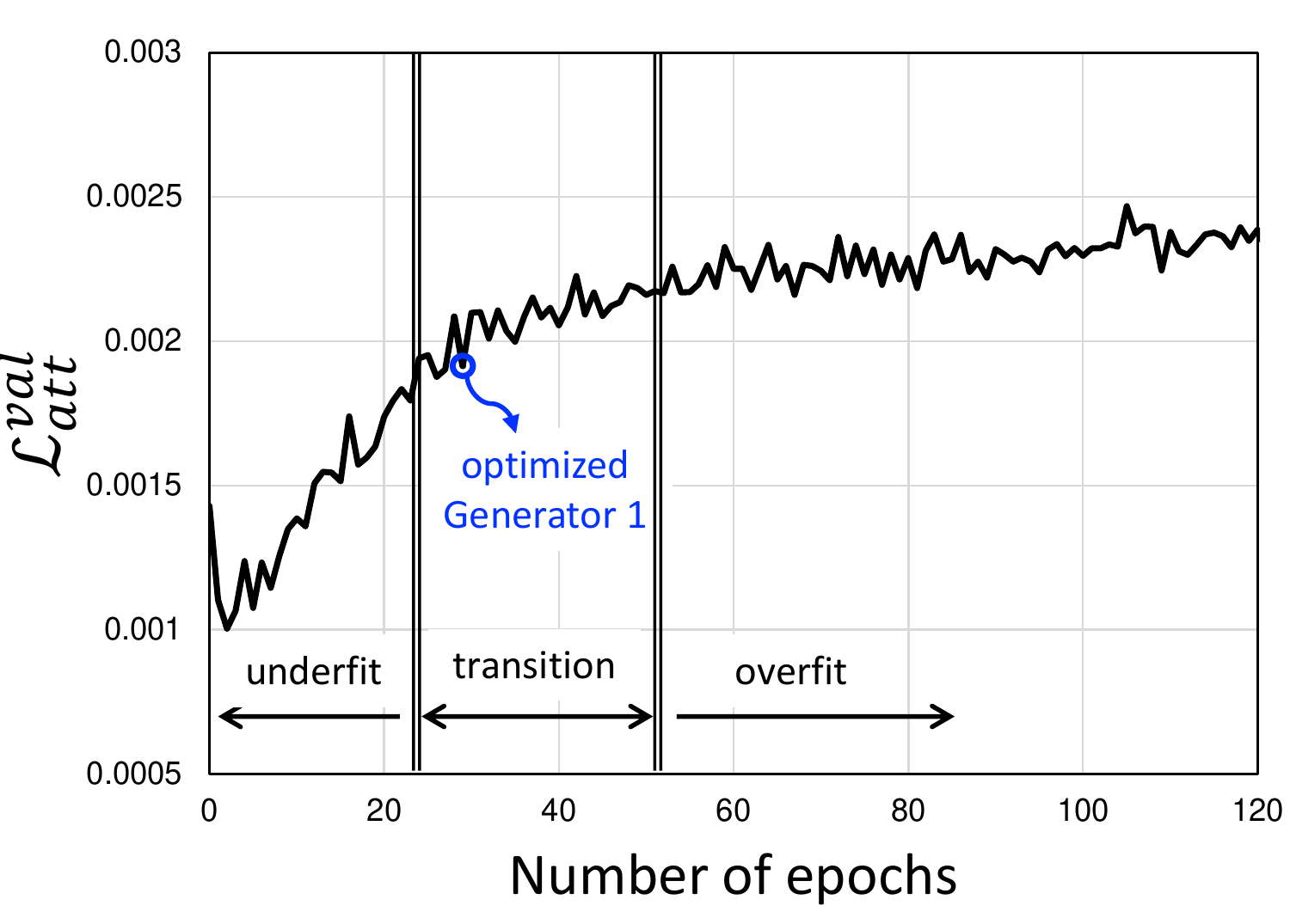}
        \caption{} \label{fig:optimized_gen2}
    \end{subfigure}
	\caption{Generator 1 training result: (a) epoch 2 predictions for the cases of $\mathcal{L}_{MAE}$ and $\mathcal{L}^{tr}_{att}$ objective functions, (i) demonstrating how the attention objective function prioritize learning the high-stress zones by initially considering a large stress for many parts, while (ii) the MAE objective function does not impose any learning priority and initially considers a uniform intermediate-stress for every part, (b) evolution of $\mathcal{L}^{val}_{att}$ with training epochs, comparing the cases of training with $\mathcal{L}_{MAE}$ and $\mathcal{L}^{tr}_{att}$ objective functions, and (c) evolution of $\mathcal{L}^{val}_{att}$ with training epochs for the case of $\mathcal{L}^{tr}_{att}$ objective function, showing different stages of training as well as the epoch leading to the optimized Generator 1.}
	\label{FIG:results}
\end{figure*}

For the case of training with the attention objective function, $\mathcal{L}^{val}_{att}$ was initially small because the objective function prioritized the learning of high-stress regions. Hence, the objective function encouraged many of the predicted pixel values to initially correspond to large magnitudes of stress, as illustrated in Figure \ref{fig:initial_epoch_result}.i. 
On the other hand, $\mathcal{L}^{val}_{att}$ was initially large for the case of training with the MAE objective function since no learning priority was imposed. In this case, the majority of the predicted pixels initially correspond to intermediate-stress values, as depicted in Figure \ref{fig:initial_epoch_result}.ii. Thus, the network gradually learned the target images with no discrimination among various parts.

Figure \ref{fig:eval_loss} depicts the evolution of $\mathcal{L}^{val}_{att}$ with the training epochs, comparing the cases of MAE and attention objective functions. As seen in the figure, $\mathcal{L}^{val}_{att}$ for the case of the attention objective function is consistently smaller than that for the case of the MAE objective function. Therefore, the proposed attention objective function could effectively improve the prediction accuracy. 

Figure \ref{fig:optimized_gen2} illustrates different training stages of Generator 1. As it can be seen, $\mathcal{L}^{val}_{att}$ initially increases at a high rate for the training epochs between 0 and 21. At this stage, the network tries to learn the generality of the target images with an emphasis on the high-stress concentration zones. For the training epochs greater than 53, the rate of increase of $\mathcal{L}^{val}_{att}$ with the training epochs becomes slower. At this stage, the network gradually overfits the training data set. Therefore, the optimized Generator 2 corresponds to an epoch in the transition state between the underfiting and overfitting training stages (i.e., between  epochs 22 and 52). As seen in Figure \ref{fig:optimized_gen2}, epoch 25, a local minimum in the transition state, is a good estimate for the epoch that optimizes Generator 2.

\subsection{Generator 2 Training Result}
\begin{figure*}
	\centering
	\includegraphics[width=0.95\textwidth]{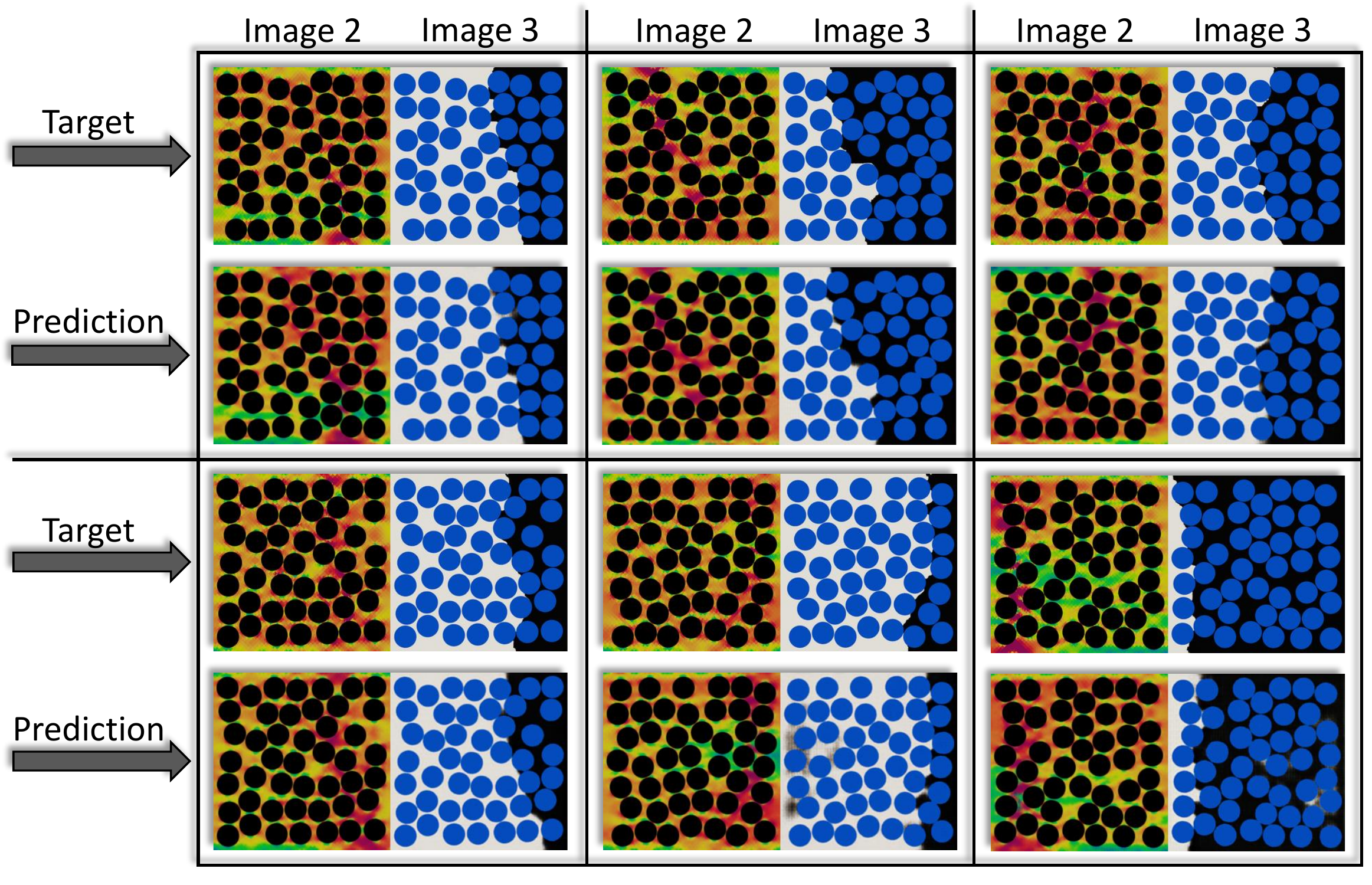}
	\caption{Sample von Mises stress at ESoDI (Image 2) and crack pattern (Image 3) predictions by the proposed deep learning framework.}
	\label{FIG:sample prediction}
\end{figure*}

In the second step, Generator 2 was trained using the outputs of the pre-trained Generator 1 as its input. After the completion of the training epochs, the accuracy of Generator 2 (i.e., the same as the overall accuracy of the deep leaning framework) was evaluated as described in Section \ref{gen 2 acc}. Depending on the objective function utilized in the training of Generator 1, the accuracy of the framework was evaluated to be
\begin{enumerate}
    \item 85\% for the case of the MAE objective function.
    \item 90\% for the case of the attention objective function.
\end{enumerate}
As a result, the proposed attention objective function could significantly improve the accuracy of the framework. These large values of accuracy demonstrate the efficacy and robustness of the proposed deep leaning framework in crack pattern prediction of CFRP composites based on the microstructural geometry. Figure \ref{FIG:sample prediction} depicts a few examples of predictions by Generator 1 and Generator 2, demonstrating the high image quality of the predicted attributes.

\subsection{Single-generator model for direct microstructure-to-crack patter translation}
In order to highlight the necessity of the intermediate prediction in the proposed framework (i.e, von Mises stress at ESoDI), a single-generator model for a direct translation of microstructure to crack pattern was tested. No intermediate prediction (i.e., von Mises stress distribution at ESoDI) was involved and the generator had to learn the significant predictors of crack pattern directly from the microstructural geometry.  The same network architecture used for Generator 1 and Generator 2 was utilized for the single-generator model. The MAE loss was used as the training objective function. The accuracy of the simplified model, based on the method described in Section \ref{gen 2 acc}, was evaluated to be 56\%. This low accuracy indicates the importance of the utilized intermediate prediction to physically guide the training towards the desired output.

\section{Conclusion}
An end-to-end deep learning framework was developed for full-field damage and failure pattern prediction in microstructure-dependent composite materials. The framework was proposed to address the downsides of the existing FE simulation approaches including complexity in constitutive equations, FE formulations, and meshing algorithms as well as the significant computational resources and run-time that they demand. Hence, being simple to use was an important consideration in designing the proposed deep learning framework. As a result, the framework was developed to work with the images of microstructures and other attributes such as displacement and stress contours that can be easily extracted from any FE simulation framework. 

A high-performance CFRP composite was used as the material of interest, and the FE simulation of several randomly generated RVEs provided the required training data set for the deep learning framework. The deep learning framework consists of two networks, namely, Generator 1 and Generator 2. Generator 1 transforms the microstructural geometry into the von Mises stress contours at ESoDI. Generator 2 translates the output of Generator 1 to the crack pattern. The proposed deep learning framework is trained sequentially; first Generator 1 is trained, and then Generator 2 is trained using the pre-trained Generator 1 outputs. In order to boost the performance, an attention objective function was designed for Generator 1 to attract the focus of the network to more important portions of the target images. 

The proposed deep learning framework resulted in a high accuracy of 90\%, evaluated based on a validation data set. By considering the fact that the materials damage and failure is one of the most complex phenomena to simulate in computational solid mechanics, the resulting high accuracy of the framework also suggests that the same approach can be effectively used to predict any other attribute at any stage of mechanical loading.

\section*{Acknowledgement}
The authors gratefully acknowledge the support from the Air-Force Office of Scientific Research (AFOSR) Young Investigator Program (YIP) award \#FA9550-20-1-0281, and Virginia Tech start-up funding. The authors also acknowledge Advanced Research Computing at Virginia Tech for providing computational resources and technical support that have contributed to the results reported within this paper. URL: https://arc.vt.edu/ 

\section*{Data availability}
Except the FE simulation framework, everything will be available on GitHub.

\appendix
\section{High-Fidelity Simulation Framework}\label{simulation framework}
In order to generating enough training data samples for the deep learning framework, several RVEs needed to be simulated. As a result, an efficient simulation framework, integrating a robust numerical method along with accurate constitutive models for different constituents was required to be used. The framework also needed to be implemented in an efficient way to speed up the computations involved in different stages of the simulation process. Hence, a finite element framework, previously developed by the authors \cite{sepasdar2020}, was employed. In the following, a summary of the finite element method and the constitutive equations, implemented within the framework is presented.

\subsection{Finite Element Method}
In this work, an IGFEM scheme was employed due to its ability to apply nonconforming mesh to complex geometries with discontinuous gradient field, such as CFRP composites \cite{soghrati2012interface,soghrati20123d}. A nonconforming mesh simplifies meshing the geometry, and most importantly, decreases the computational burden by enabling the application of a uniform mesh to the geometry. Nonlinear cohesive IGFEM is an extension of IGFEM that allows an element to include a number of arbitrary oriented embedded cohesive interfaces, modeled by CZMs. The portions of an element traversed by interfaces can have different constitutive behaviors. For more details on the formulation and implementation of nonlinear cohesive IGFEM, the reader can refer to \cite{shakiba2019transverse,Zacek2020,sepasdar2020}. The described method was implemented within an IGFEM solver, developed using the C++ language. The IGFEM framework was designed to be fast by the virtue of object oriented programming, efficient memory management, and utilizing powerful scientific libraries (e.g., MPI, PETSC, GSL, METIS) to run the simulations fully in parallel. 

In the framework, three-node triangular plane-strain elements were used that can contain up to two enriched interfaces. A mesh sensitivity analysis determined a uniform mesh size such that 20 elements to be along the diameters of each fiber. A strain of 0.02 was applied to the RVEs right edge incrementally, while the horizontal displacement of the left edge was restrained. This boundary condition subjects the RVE to uniaxial transverse tension while allows for free vertical deformations. 

\subsection{Constitutive Equations}
Transverse cracks in CFRP composites are the result of fiber/matrix interface debondings, accompanied by matrix cracking in between the adjacent fibers. The crack does not pass through fibers, and fiber breakage rarely occurs \cite{mortell2014situ,montgomery2018multiscale}. Therefore, in a robust numerical model, accurate nonlinear constitutive behaviors for matrix and fiber/matrix interfaces must be considered while the fibers can be assumed elastic. The constitutive models summarized here were previously implemented and verified by the authors \cite{sepasdar2020}.

In this work, the matrix is modeled using an elasto-plastic damage model with a brittle failure. The initiation of plastic deformation is specified by the Tschoegl yield criterion \cite{tschoegl1971failure} as
\begin{eqnarray}\label{eq:1}
\phi ( \pmb{\upsigma}) = 6J_2 + 2I_1(\sigma^c_y - \sigma^t_y) - 2\sigma^c_y \sigma^t_y
\end{eqnarray}
where $\pmb{\upsigma}$ is the stress tensor, $I_1$ and $J_1$ are the first and second invariants of the stress and deviatoric stress tensors, respectively, and $\sigma^t_y$ and $\sigma^c_y$ are the matrix strengths under uniaxial tension and compression, respectively. The evolution of plastic deformation is modeled by an isotropic hardening law, defined using a non-associated flow rule, based on the von Mises criterion. To provide a smooth elastic-to-plastic transition, a behavior proposed by Ramberg and Osgood \cite{ramberg1943description} was considered in the constitutive model. The behavior is defined as
\begin{equation} \label{eq:3}
    H=a\left(\frac{\sigma_Y}{\sigma_{von}}\right)^b
\end{equation}
where $H$ is the slope of the tangent line to the plastic branch, $a$ and $b$ are two parameters that control the shape and smoothness of the elastic-to-plastic transition, and $\sigma_Y$ and $\sigma_{von}$ are the von Mises stresses at initiation and during the evolution of plastic deformation, respectively.

The initiation of failure is determined by the Tschoegl yield criterion \cite{tschoegl1971failure} formulated based on the strain tensor, $\pmb{\upepsilon}$, as
\begin{eqnarray}\label{eq:2}
\phi' ( \pmb{\upepsilon}) = 6J'_2 + 2I'_1(\epsilon^c - \epsilon^t) - 2\epsilon^c \epsilon^t
\end{eqnarray}
where $I'_1$ and $J'_1$ are the first and second invariants of the strain and deviatoric strain tensors, respectively, and $\epsilon^t_y$ and $\epsilon^c_y$ are the failure strains under uniaxial tension and compression, respectively. After the failure initiates, the evolution of
damage is modeled by the Simo and Ju continuum damage constitutive equation \cite{simo1987strain,simo1989continuum}. In the constitutive behavior, the damaged stiffness matrix, $\mathbf D^{d}$, is calculated by the penalization of the elasto-plastic stiffness matrix, $\mathbf D^{ep}$, via a damage index, $d$, as $\mathbf D^d = (1-d)\mathbf D^{ep}$. The damage index $d$ is calculated and updated via
\begin{equation} \label{eq:damage evolution}
    d= d + \left(\frac{d t}{1+\mu d t}\right)(G-Y)
\end{equation}
where $\mu$ is a viscosity parameter, $dt$ is the pseudo-time, $G$ is the damage parameter, and $Y$ is the damage threshold. $G$ and $Y$ are defined as 
\begin{equation} \label{eq:damage function}
    G=1-\bar{\tau_0}\frac{1-A}{\bar{\tau}}-A\exp(B(\bar{\tau_0}-\bar{\tau}))
\end{equation}
\begin{equation} \label{eq:damage threshold}
    Y=\frac{Y+\mu d t G}{1+\mu d t}
\end{equation}
where $A$ and $B$ are two damage parameter constants, $\bar{\tau}=\sqrt{2\Xi}$ is a damage parameter calculated based on the strain energy $\Xi$, and $\bar{\tau_0}$ is a constant called the initial damage threshold which is equal to $\bar{\tau}$ at the start of damage. The damage index $d$ is updated only when $G>0$, $Y>0$, and $G-Y>0$. 

The debonding of fiber matrix interfaces is simulated using a CZM constitutive model proposed by Ortiz and Pandolfi \cite{ortiz1999finite}. In the model, the cohesive behavior is defined based on an effective opening displacement, $\delta=\sqrt{\delta_n^2+\delta_t^2}$, taking into account the interaction between normal and tangential debonding displacements (i.e., $\delta_n$ and $\delta_t$, respectively). The bilinear traction-separation behavior illustrated in Figure \ref{FIG:CZM} was assumed for the interfaces. The cohesive parameters in the behavior are cohesive strength, $T_c$, critical opening displacement, $\delta_c$, and fracture toughness, $G_c$. CZMs can cause convergence issue in static analyses when modeling brittle debonding failures. The convergence issue is caused by the Newton-Raphson iterations entering an infinite cycle \cite{sepasdar2020overcoming}. Therefore, an artificial viscosity proposed by \cite{gao2004simple} was added in the cohesive law to overcome the difficulty.

\thefigure
\setcounter{figure}{0}  
\begin{figure*}
	\centering
	\includegraphics[width=0.6\textwidth]{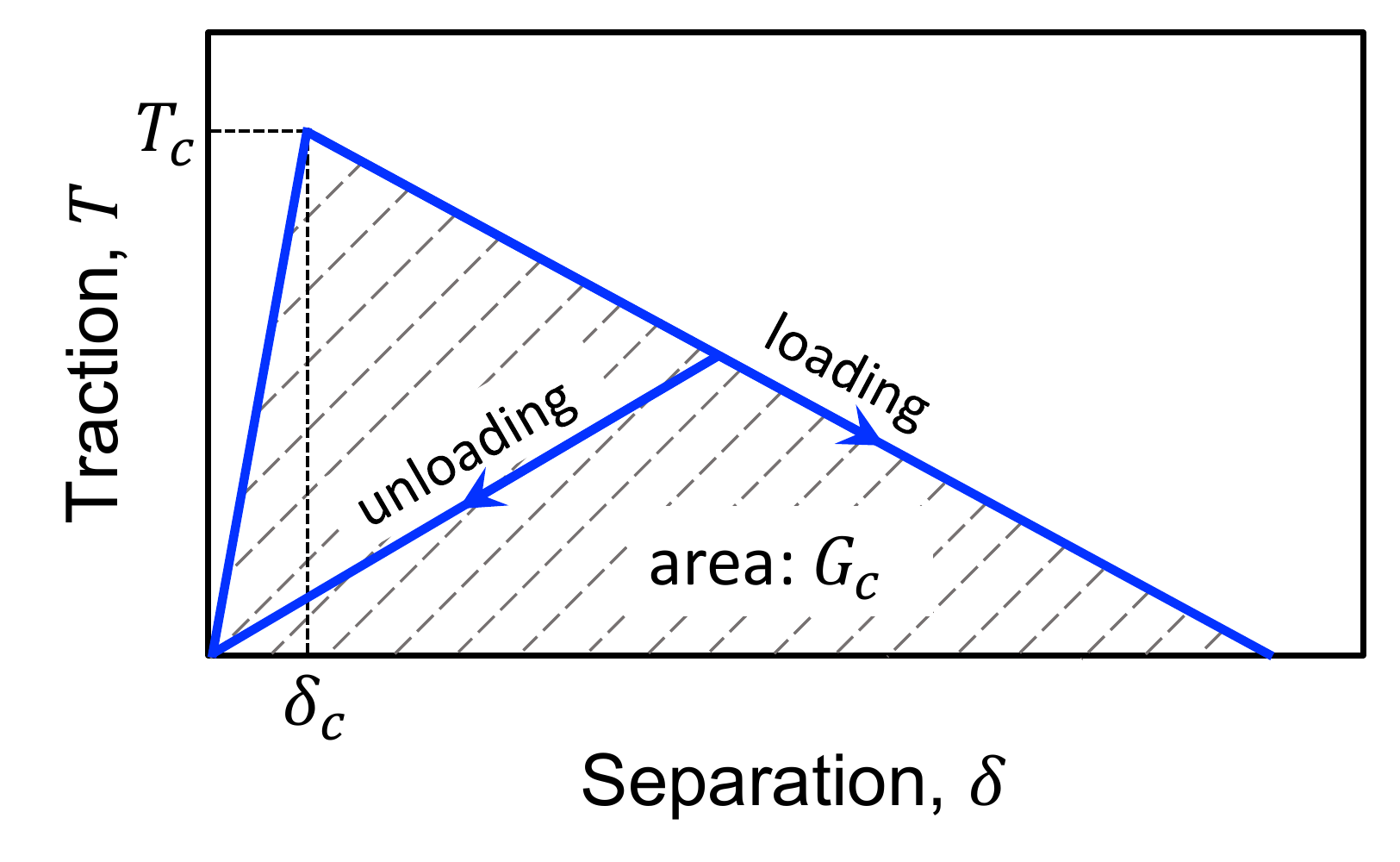}
	\caption{The bilinear traction-separation constitutive law used for modeling fiber/matrix cohesive interfaces.}
	\label{FIG:CZM}
\end{figure*}

The parameters of the aforementioned constitutive models for different components of a typical CFRP composite is presented in Table \ref{Table:material properties}. For more information regarding the constitutive models' derivation, implementation, and calibration as well as the verification of the developed IGFEM framework for the transverse failure simulation of CFRP composites, the reader is referred to \cite{sepasdar2020}. 

The FE simulations of the 4500 generated RVEs were conducted using the Dragon's Tooth cluster at Advanced Research Computing (ARC) at Virginia Tech. 24 Intel Xeon E5-2680v3 (Haswell) 2.5 GHz processors were requested and the run-time to simulate all the RVEs was 9 days.

\begin{table}[]
    \centering
    \caption{Material properties for different constituents of the CFRP composite. $E$, $G$, and $\nu$ are Young's modulus, the shear modulus, and the Poisson's ratio, respectively. 1 and 2 represent the out-of-plane and in-plane directions, respectively, and 3 is the direction perpendicular to the 1 and 2 directions.} \label{Table:material properties}
\begin{tabular}{clclc}
\hline
\multicolumn{3}{|c|}{Matrix}                                                     & \multicolumn{2}{c|}{Carbon fiber}            \\ \hline
\multicolumn{1}{|c|}{Elastic}    & $E (GPa)$        & \multicolumn{1}{c|}{3.9}   & $E_1 (MPa)$      & \multicolumn{1}{c|}{233}  \\
\multicolumn{1}{|c|}{properties} & $\nu$            & \multicolumn{1}{c|}{0.39}  & $E_2 (MPa)$      & \multicolumn{1}{c|}{23.1} \\ \cline{1-1}
\multicolumn{1}{|c|}{}           & $\sigma^c (MPa)$ & \multicolumn{1}{c|}{79}    & $G_{12} (MPa)$   & \multicolumn{1}{c|}{8.96} \\
\multicolumn{1}{|c|}{Plastic}    & $\sigma^t (MPa)$ & \multicolumn{1}{c|}{62}    & $G_{23} (MPa)$   & \multicolumn{1}{c|}{8.27} \\
\multicolumn{1}{|c|}{properties} & $H$              & \multicolumn{1}{c|}{20000} & $\nu_{12}$       & \multicolumn{1}{c|}{0.2}  \\
\multicolumn{1}{|c|}{}           & $n$              & \multicolumn{1}{c|}{12}    &                  & \multicolumn{1}{c|}{}     \\ \cline{1-1} \cline{4-5} 
\multicolumn{1}{|c|}{}           & $\epsilon^c$     & \multicolumn{1}{c|}{0.35}  & \multicolumn{2}{c|}{Fiber/matrix}            \\
\multicolumn{1}{|c|}{Damage}     & $\epsilon^t$     & \multicolumn{1}{c|}{0.04}  & \multicolumn{2}{c|}{cohesive interfaces}     \\ \cline{4-5} 
\multicolumn{1}{|c|}{properties} & $A$              & \multicolumn{1}{c|}{0.95}  & $T_c (MPa)$ & \multicolumn{1}{c|}{70}   \\
\multicolumn{1}{|c|}{}           & $B$              & \multicolumn{1}{c|}{2}     & $delta_c (nm)$   & \multicolumn{1}{c|}{1}    \\
\multicolumn{1}{|c|}{}           & $\mu$            & \multicolumn{1}{c|}{10}    & $G_c (N/m)$      & \multicolumn{1}{c|}{8.75} \\ \hline
\multicolumn{1}{l}{}             &                  & \multicolumn{1}{l}{}       &                  &                          
\end{tabular}
\end{table}

\bibliography{Paper}

\begin{thebibliography}{10}
\expandafter\ifx\csname url\endcsname\relax
  \def\url#1{\texttt{#1}}\fi
\expandafter\ifx\csname urlprefix\endcsname\relax\def\urlprefix{URL }\fi
\expandafter\ifx\csname href\endcsname\relax
  \def\href#1#2{#2} \def\path#1{#1}\fi

\bibitem{raissi2019deep}
M.~Raissi, Z.~Wang, M.~S. Triantafyllou, G.~E. Karniadakis, Deep learning of
  vortex-induced vibrations, Journal of Fluid Mechanics 861 (2019) 119--137.

\bibitem{muralidhar2020physics}
N.~Muralidhar, J.~Bu, Z.~Cao, L.~He, N.~Ramakrishnan, D.~Tafti, A.~Karpatne,
  Physics-guided deep learning for drag force prediction in dense
  fluid-particulate systems, Big Data 8~(5) (2020) 431--449.

\bibitem{raissi2020hidden}
M.~Raissi, A.~Yazdani, G.~E. Karniadakis, Hidden fluid mechanics: Learning
  velocity and pressure fields from flow visualizations, Science 367~(6481)
  (2020) 1026--1030.

\bibitem{kashefi2021point}
A.~Kashefi, D.~Rempe, L.~J. Guibas, A point-cloud deep learning framework for
  prediction of fluid flow fields on irregular geometries, Physics of Fluids
  33~(2) (2021) 027104.

\bibitem{cecen2018material}
A.~Cecen, H.~Dai, Y.~C. Yabansu, S.~R. Kalidindi, L.~Song, Material
  structure-property linkages using three-dimensional convolutional neural
  networks, Acta Materialia 146 (2018) 76--84.

\bibitem{yang2018deep}
Z.~Yang, Y.~C. Yabansu, R.~Al-Bahrani, W.-k. Liao, A.~N. Choudhary, S.~R.
  Kalidindi, A.~Agrawal, Deep learning approaches for mining structure-property
  linkages in high contrast composites from simulation datasets, Computational
  Materials Science 151 (2018) 278--287.

\bibitem{vlassis2020geometric}
N.~N. Vlassis, R.~Ma, W.~Sun, Geometric deep learning for computational
  mechanics part i: Anisotropic hyperelasticity, Computer Methods in Applied
  Mechanics and Engineering 371 (2020) 113299.

\bibitem{herriott2020predicting}
C.~Herriott, A.~D. Spear, Predicting microstructure-dependent mechanical
  properties in additively manufactured metals with machine-and deep-learning
  methods, Computational Materials Science 175 (2020) 109599.

\bibitem{wang2018multiscale}
K.~Wang, W.~Sun, A multiscale multi-permeability poroplasticity model linked by
  recursive homogenizations and deep learning, Computer Methods in Applied
  Mechanics and Engineering 334 (2018) 337--380.

\bibitem{mozaffar2019deep}
M.~Mozaffar, R.~Bostanabad, W.~Chen, K.~Ehmann, J.~Cao, M.~Bessa, Deep learning
  predicts path-dependent plasticity, Proceedings of the National Academy of
  Sciences 116~(52) (2019) 26414--26420.

\bibitem{heider2020so}
Y.~Heider, K.~Wang, W.~Sun, So (3)-invariance of informed-graph-based deep
  neural network for anisotropic elastoplastic materials, Computer Methods in
  Applied Mechanics and Engineering 363 (2020) 112875.

\bibitem{donegan2019associating}
S.~P. Donegan, N.~Kumar, M.~A. Groeber, Associating local microstructure with
  predicted thermally-induced stress hotspots using convolutional neural
  networks, Materials Characterization 158 (2019) 109960.

\bibitem{feng2020difference}
H.~Feng, P.~Prabhakar, Difference-based deep learning framework for stress
  predictions in heterogeneous media, arXiv preprint arXiv:2007.04898.

\bibitem{zhang2020physics}
E.~Zhang, M.~Yin, G.~E. Karniadakis, Physics-informed neural networks for
  nonhomogeneous material identification in elasticity imaging, arXiv preprint
  arXiv:2009.04525.

\bibitem{abueidda2020deep}
D.~W. Abueidda, Q.~Lu, S.~Koric, Deep learning collocation method for solid
  mechanics: Linear elasticity, hyperelasticity, and plasticity as examples,
  arXiv preprint arXiv:2012.01547.

\bibitem{haghighat2021sciann}
E.~Haghighat, R.~Juanes, Sciann: A keras/tensorflow wrapper for scientific
  computations and physics-informed deep learning using artificial neural
  networks, Computer Methods in Applied Mechanics and Engineering 373 (2021)
  113552.

\bibitem{dong2020microstructural}
Y.~Dong, C.~Su, P.~Qiao, L.~Sun, Microstructural crack segmentation of
  three-dimensional concrete images based on deep convolutional neural
  networks, Construction and Building Materials 253 (2020) 119185.

\bibitem{pazdernik2020microstructural}
K.~Pazdernik, N.~L. LaHaye, C.~M. Artman, Y.~Zhu, Microstructural
  classification of unirradiated lialo2 pellets by deep learning methods,
  Computational Materials Science 181 (2020) 109728.

\bibitem{lei2020lost}
X.~Lei, L.~Sun, Y.~Xia, Lost data reconstruction for structural health
  monitoring using deep convolutional generative adversarial networks,
  Structural Health Monitoring (2020) 1475921720959226.

\bibitem{mao2020toward}
J.~Mao, H.~Wang, B.~F. Spencer~Jr, Toward data anomaly detection for automated
  structural health monitoring: Exploiting generative adversarial nets and
  autoencoders, Structural Health Monitoring (2020) 1475921720924601.

\bibitem{soleimani2021system}
M.~H. Soleimani-Babakamali, R.~Sepasdar, K.~Nasrollahzadeh, R.~Sarlo,
  System-reliability based multi-ensemble of gan and one-class joint gaussian
  distributions for unsupervised real-time structural health monitoring, arXiv
  preprint arXiv:2102.01158.

\bibitem{hunter2019reduced}
A.~Hunter, B.~A. Moore, M.~Mudunuru, V.~Chau, R.~Tchoua, C.~Nyshadham,
  S.~Karra, D.~O’Malley, E.~Rougier, H.~Viswanathan, et~al., Reduced-order
  modeling through machine learning and graph-theoretic approaches for brittle
  fracture applications, Computational Materials Science 157 (2019) 87--98.

\bibitem{schwarzer2019learning}
M.~Schwarzer, B.~Rogan, Y.~Ruan, Z.~Song, D.~Y. Lee, A.~G. Percus, V.~T. Chau,
  B.~A. Moore, E.~Rougier, H.~S. Viswanathan, et~al., Learning to fail:
  Predicting fracture evolution in brittle material models using recurrent
  graph convolutional neural networks, Computational Materials Science 162
  (2019) 322--332.

\bibitem{pierson2019predicting}
K.~Pierson, A.~Rahman, A.~D. Spear, Predicting microstructure-sensitive
  fatigue-crack path in 3d using a machine learning framework, JOM 71~(8)
  (2019) 2680--2694.

\bibitem{gamstedt1999micromechanisms}
E.~Gamstedt, B.~Sj{\"o}gren, Micromechanisms in tension-compression fatigue of
  composite laminates containing transverse plies, Composites Science and
  Technology 59~(2) (1999) 167--178.

\bibitem{hobbiebrunken2006evaluation}
T.~Hobbiebrunken, M.~Hojo, T.~Adachi, C.~De~Jong, B.~Fiedler, Evaluation of
  interfacial strength in {CF}/epoxies using {FEM} and in-situ experiments,
  Composites Part A: Applied Science and Manufacturing 37~(12) (2006)
  2248--2256.

\bibitem{montgomery2018multiscale}
C.~B. Montgomery, Multiscale characterization of carbon fiber-reinforced epoxy
  composites, Ph.D. thesis, University of Illinois at Urbana-Champaign (2018).

\bibitem{Hernandez2020}
L.~Hernandez, R.~Sepasdar, M.~Shakiba, Sensitivity of crack formation in
  fiber-reinforced composites to microstructural geometry and interfacial
  properties, in: Proceeding of the American Society for Composites,
  Thirty-Fifth Technical Conference, DEStech Publications, Inc., 2020, pp.
  1576--1591.
\newblock \href {http://dx.doi.org/10.12783/asc35/34954}
  {\path{doi:10.12783/asc35/34954}}.

\bibitem{sepasdar2020}
R.~Sepasdar, M.~Shakiba, Micromechanical study of multiple transverse cracking
  in cross-ply fiber-reinforced composite laminates.

\bibitem{zacek2017exploring}
S.~A. Zacek, Exploring the link between microstructure statistics and
  transverse ply fracture in carbon/epoxy composites.

\bibitem{ronneberger2015u}
O.~Ronneberger, P.~Fischer, T.~Brox, U-net: Convolutional networks for
  biomedical image segmentation, in: International Conference on Medical image
  computing and computer-assisted intervention, Springer, 2015, pp. 234--241.

\bibitem{zhou2018unet++}
Z.~Zhou, M.~M.~R. Siddiquee, N.~Tajbakhsh, J.~Liang, Unet++: A nested u-net
  architecture for medical image segmentation, in: Deep learning in medical
  image analysis and multimodal learning for clinical decision support,
  Springer, 2018, pp. 3--11.

\bibitem{ibtehaz2020multiresunet}
N.~Ibtehaz, M.~S. Rahman, Multiresunet: Rethinking the u-net architecture for
  multimodal biomedical image segmentation, Neural Networks 121 (2020) 74--87.

\bibitem{koeppe2020intelligent}
A.~Koeppe, F.~Bamer, B.~Markert, An intelligent nonlinear meta element for
  elastoplastic continua: deep learning using a new time-distributed residual
  u-net architecture, Computer Methods in Applied Mechanics and Engineering 366
  (2020) 113088.

\bibitem{tang2021deep}
M.~Tang, Y.~Liu, L.~J. Durlofsky, Deep-learning-based surrogate flow modeling
  and geological parameterization for data assimilation in 3d subsurface flow,
  Computer Methods in Applied Mechanics and Engineering 376 (2021) 113636.

\bibitem{soghrati2012interface}
S.~Soghrati, A.~M. Arag{\'o}n, C.~Armando~Duarte, P.~H. Geubelle, An
  interface-enriched generalized {FEM} for problems with discontinuous gradient
  fields, International Journal for Numerical Methods in Engineering 89~(8)
  (2012) 991--1008.

\bibitem{soghrati20123d}
S.~Soghrati, P.~H. Geubelle, A {3D} interface-enriched generalized finite
  element method for weakly discontinuous problems with complex internal
  geometries, Computer Methods in Applied Mechanics and Engineering 217 (2012)
  46--57.

\bibitem{shakiba2019transverse}
M.~Shakiba, D.~R. Brandyberry, S.~Zacek, P.~H. Geubelle, Transverse failure of
  carbon fiber composites: Analytical sensitivity to the distribution of
  fiber/matrix interface properties, International Journal for Numerical
  Methods in Engineering 120~(5) (2019) 650--665.

\bibitem{Zacek2020}
S.~Zacek, D.~Brandyberry, A.~Klepacki, C.~Montgomery, M.~Shakiba, M.~Rossol,
  A.~Najafi, X.~Zhang, N.~Sottos, P.~Geubelle, C.~Przybyla, G.~Jefferson,
  Transverse failure of unidirectional composites: Sensitivity to interfacial
  properties, Springer International Publishing, Cham, 2020, pp. 329--347.

\bibitem{mortell2014situ}
D.~Mortell, D.~Tanner, C.~McCarthy, In-situ {SEM} study of transverse cracking
  and delamination in laminated composite materials, Composites Science and
  Technology 105 (2014) 118--126.

\bibitem{tschoegl1971failure}
N.~Tschoegl, Failure surfaces in principal stress space, in: Journal of polymer
  science Part C: Polymer symposia, Vol.~32, Wiley Online Library, 1971, pp.
  239--267.

\bibitem{ramberg1943description}
W.~Ramberg, W.~R. Osgood, Description of stress-strain curves by three
  parameters.

\bibitem{simo1987strain}
J.~C. Simo, J.~Ju, Strain-and stress-based continuum damage models—{I}.
  formulation, International journal of solids and structures 23~(7) (1987)
  821--840.

\bibitem{simo1989continuum}
J.~Simo, J.~Ju, On continuum damage-elastoplasticity at finite strains,
  Computational Mechanics 5~(5) (1989) 375--400.

\bibitem{ortiz1999finite}
M.~Ortiz, A.~Pandolfi, Finite-deformation irreversible cohesive elements for
  three-dimensional crack-propagation analysis, International Journal for
  Numerical Methods in Engineering 44~(9) (1999) 1267--1282.

\bibitem{sepasdar2020overcoming}
R.~Sepasdar, M.~Shakiba, Overcoming the convergence difficulty of cohesive zone
  models through a newton-raphson modification technique, Engineering Fracture
  Mechanics (2020) 107046.~\href
  {http://dx.doi.org/10.1016/j.engfracmech.2020.107046}
  {\path{doi:10.1016/j.engfracmech.2020.107046}}.

\bibitem{gao2004simple}
Y.~Gao, A.~Bower, A simple technique for avoiding convergence problems in
  finite element simulations of crack nucleation and growth on cohesive
  interfaces, Modelling and Simulation in Materials Science and Engineering
  12~(3) (2004) 453.

\end{thebibliography}
\end{document}